\documentclass[runningheads]{llncs}
\usepackage[T1]{fontenc}
\usepackage{graphicx}
\usepackage{xcolor}
\usepackage{comment}
\usepackage{tabularx}
\usepackage{float}
\usepackage{url}
\usepackage{hyperref}
\usepackage{listings}
\newcolumntype{L}[1]{>{\raggedright\let\newline\\\arraybackslash\hspace{0pt}}m{#1}}
\newcolumntype{C}[1]{>{\centering\let\newline\\\arraybackslash\hspace{0pt}}m{#1}}
\newcolumntype{R}[1]{>{\raggedleft\let\newline\\\arraybackslash\hspace{0pt}}m{#1}}

\begin{document}

\title{Gypscie: A Cross-Platform AI Artifact Management System}

\author{
Fabio Porto\inst{1} \and
Eduardo Ogasawara\inst{2} \and
Gabriela Moraes Botaro\inst{1} \and
Julia Neumann Bastos\inst{1} \and
Augusto Fonseca\inst{1} \and
Esther Pacitti\inst{3} \and
Patrick Valduriez\inst{1,3} 
}

\authorrunning{F. Porto et al.}

\institute{
LNCC, Petrópolis, Brazil \and
CEFET/RJ, Rio de Janeiro, Brazil \and
Inria, Univ Montpellier, CNRS, LIRMM, France
}

\maketitle

\begin{abstract}
Artificial Intelligence (AI) models, encompassing both traditional machine learning (ML) and more advanced approaches such as deep learning and large language models (LLMs), play a central role in modern applications. AI model lifecycle management involves the end-to-end process of managing these models, from data collection and preparation to model building, evaluation, deployment, and continuous monitoring. This process is inherently complex, as it requires the coordination of diverse services that manage AI artifacts such as datasets, dataflows, and models, all orchestrated to operate seamlessly. In this context, it is essential to isolate applications from the complexity of interacting with heterogeneous services, datasets, and AI platforms.

In this paper,
we introduce Gypscie, a cross-platform AI artifact management system. By providing a unified view of all AI artifacts, Gypscie simplifies the development and deployment of AI applications. This unified view is realized through a knowledge graph that captures application semantics and a rule-based query language that supports reasoning over data and models. Model lifecycle activities are represented as high-level dataflows that can be scheduled across multiple platforms, such as servers, cloud platforms, or supercomputers. Finally, Gypscie records provenance information about the artifacts it produces, thereby enabling explainability.
Our qualitative comparison with representative AI systems shows that Gypscie supports a broader range of functionalities across the AI artifact lifecycle. Our experimental evaluation demonstrates that Gypscie can successfully optimize and schedule dataflows on AI platforms from an abstract specification.

\end{abstract}

\section{Introduction}

In modern scientific and operational applications, AI models play a critical role and introduce new challenges for application development and execution. These models include both traditional machine learning (ML) models and more advanced approaches such as deep learning models and large language models (LLMs). Consequently, AI model lifecycle management \cite{ibm_ai_2020}, defined as the process of building, deploying, and maintaining AI models over time, has become crucial for reducing the complexity of handling different model services, heterogeneous datasets, and multiple AI platforms.

The AI model lifecycle, often referred to in the literature as the ML model lifecycle \cite{schlegel_management_2023}, typically involves four key stages:
\begin{enumerate}
\item \textbf{Requirements stage}: derives the requirements for the model based on the application needs, identifies relevant data sources, and determines the appropriate learning strategy.
\item \textbf{Data-oriented stage}: involves collecting, cleaning, preprocessing, and preparing the data for training the ML model.
\item \textbf{Model-oriented stage}: selects the appropriate AI learning algorithm for the intended task, tunes its hyperparameters, trains it using the preprocessed data, and evaluates its performance according to selected metrics. For LLMs, evaluation also involves assessing whether outputs align with ethical guidelines and are free from toxic content.
\item \textbf{Operations stage}: deploys the trained model into a production environment, configures it with specific software libraries, and manages the execution strategy, such as centralized or distributed, on demand or continuous. During production, model performance is monitored and flagged as outdated when necessary.
\end{enumerate}

During these stages, a wide range of artifacts is produced. Typical ML artifacts include datasets, learners, trained models, model versions, feature sets, dataflows (e.g., ML pipelines), and user functions that provide the desired behavior for an operation. Additionally, provenance metadata management records both prospective and retrospective information about the use and generation of artifacts. LLMs share core ML artifacts but add new ones reflecting their scale, multi-stage training, and prompt-driven behavior \cite{clearml_clearml_2025}. In addition to massive pretraining corpora and tokenized datasets, they include specialized instruction-tuning and Reinforcement Learning with Human Feedback (RLHF) datasets, as well as benchmark collections. Model artifacts include not only base and fine-tuned weights but also distilled models and Mixture-of-Experts variants.

Thus, AI artifacts are essential to support AI model lifecycle management, particularly to achieve comparability, traceability, and reproducibility of model and data artifacts across all lifecycle stages \cite{schlegel_management_2023}. However, managing the wide range of services involved in the artifact lifecycle is a complex task, which requires orchestrating services across different stages, each operating on distinct artifacts. The execution of these services must be continuously monitored with evaluation and analytical tools. For example, real-time monitoring is essential to provide online feedback on training metrics and inference quality. Dataset preparation also poses challenges, as it involves building and executing preprocessing dataflows, handling heterogeneous dataset formats, and selecting appropriate transformation functions \cite{bala_brief_2024}. Moreover, the AI model lifecycle is inherently cyclic, since training data can become outdated or concept drift can be detected \cite{hoens_learning_2012}. In such cases, new versions of both data and models must be generated, which demands systematic management of multiple artifact versions within a continuous process.

A model’s lifecycle generates heterogeneous data that must be systematically managed. First, metadata describe and contextualize the lifecycle of artifacts, including details about the learner used for training, computed metrics, and the datasets employed. These must be stored, often in a relational database, and supported by retrieval and interpretation interfaces. Second, training and inference datasets form another critical class of data. Their types vary with the application, from time series collected through real-time sensors to tabular data, images, graphs, or token sequences for LLMs. Managed datasets may also enrich predictions with domain-specific context, such as vegetation type and altitude in urban environments. Third, to enable explainability, provenance data are captured and stored, often through dedicated mechanisms \cite{davison_automated_2012} and graph databases \cite{ramusat_efficient_2022}. Finally, predictions themselves constitute another dataset requiring management. Handling and extracting value from this wide spectrum of heterogeneous data remains a major challenge.

Furthermore, an application may rely on several candidate models to compose a prediction, for example, models based on different architectures trained on the same or similar datasets. The application must identify the appropriate candidate models for a given task, invoke them, and interpret the resulting predictions.

Some activities may also impose specific demands on the execution platform. For instance, when experimenting with different learners and a small dataset \cite{da_silva_conceptual_2019}, a local system can simplify the training process. When real training data are used, a deep learning model may require a platform with GPUs and sufficient VRAM. For a large-scale model with billions of parameters, the cloud or a supercomputer may become the preferred AI platform. For ML inference, scaling according to the number of concurrent requests also determines the size and characteristics of the AI platform.

The manual management of AI artifacts within the AI application itself is inefficient, time-consuming, and highly complex. Therefore, AI artifact management systems have been proposed \cite{schlegel_management_2023} to support the systematic collection, storage, correlation, and management of AI lifecycle artifacts, making them findable, reusable, interpretable, version-controlled, and seamlessly integrated with lifecycle services. However, these systems vary considerably in scope. Many remain specialized in particular lifecycle stages, artifact types, or execution ecosystems, which makes it difficult to obtain an integrated view of artifacts and their relationships across the complete AI lifecycle.

In contrast, Gypscie is a cross-platform AI artifact management system that unifies AI artifact management across the entire AI model lifecycle. It was developed in the context of scientific domains such as meteorology, drug discovery, and oil and gas exploration. The term Gypscie is a blend of ``gypsy'' (prediction) and ``science'', conveying the idea that the system supports predictions for scientific applications. Gypscie integrates heterogeneous AI platforms across on-premise servers, clusters, and HPC environments.

In this paper, we introduce Gypscie and its open, service-based architecture. Gypscie is both a system, providing services for AI artifact management, and a platform that others can build on or extend. Gypscie simplifies the development and deployment of AI applications by providing a unified view of all AI artifacts. This view is supported by a knowledge graph that captures the application's semantics and a rule-based query language that allows reasoning about artifacts. Dataflows are high-level representations of data pipelines that move and transform data, typically for training, validation, and deployment of AI models. They can be scheduled to run as jobs on a target AI platform, for instance, as a Scikit-learn processing job on a server, an interactive job on a Spark cluster, or a batch job on a supercomputer. Gypscie keeps track of provenance information about the resulting dataflow artifacts, enabling explainability. Finally, by enabling the sharing of different kinds of artifacts, Gypscie fosters scientific collaboration.

This paper addresses two main questions: how AI artifacts can be managed in an integrated way across the AI lifecycle in heterogeneous AI platforms, and how abstract AI dataflows can be represented and instantiated across heterogeneous platforms while preserving semantic consistency and enabling optimization.

This paper makes three main contributions. First, it presents a cross-platform architecture for integrated AI artifact management across the lifecycle, covering datasets, models, functions, dataflows, and provenance. Second, it introduces a knowledge-graph-based integrated view, supported by a rule-based query language, to connect artifacts, metadata, domain data, and provenance information. Third, it proposes a high-level dataflow model with cross-platform instantiation and scheduling support, and evaluates this functionality experimentally while assessing the system more broadly through qualitative comparison.

The rest of this paper is organized as follows. Section \ref{sec_example} presents a motivating example from a real-life scientific application. Section \ref{sec_architecture} gives the system overview of Gypscie. Sections \ref{sec_Modelmanagement}, \ref{sec_knowledge-graph}, and \ref{sec_dataflow} discuss three major capabilities of Gypscie: model management, knowledge graph, and dataflow processing, respectively. Section \ref{sec_evaluation} provides an evaluation of the Gypscie system.
Section \ref{sec_related-work} reviews related work. Finally, Section \ref{sec_conclusion} concludes the paper.

\section{Motivating Example}
\label{sec_example}

To illustrate the challenges of AI artifact management in scientific applications, we present a motivating example drawn from the prediction of extreme weather events. This use case also serves as the basis for validating our approach, as discussed in Section \ref{sec_evaluation}.

Extreme rainfall events are occurring with increasing frequency and intensity worldwide. In Rio de Janeiro, such events have a severe impact every year, including traffic disruptions, landslides, property damage, and even loss of life. To mitigate these impacts, the city of Rio de Janeiro relies on its operations center, the Centro de Operações Rio (COR)\footnote{\url{https://cor.rio}}. COR functions as a citywide data hub, aggregating information on factors such as traffic congestion and weather conditions to keep both government officials and citizens informed. It is also responsible for continuous weather monitoring, operating 24 hours a day, seven days a week.

The Rionowcast project at COR, in which we are involved, develops and operates AI models to forecast rainfall up to six hours in advance. However, precise rainfall prediction remains challenging because of the chaotic nature of atmospheric phenomena. To address this, the project leverages a wide range of weather data sources, including radar, rain gauges, ocean buoys, radiosondes, and weather stations. Project activities are carried out by data engineers, model developers, and domain scientists, such as meteorologists.

Data engineers manage the lifecycle of data-based artifacts. They use historical weather records to build training datasets, retaining only days with rainfall above a given threshold. Streaming data are continuously captured and fed into the models for inference. Each data source passes through a preprocessing dataflow that cleans, normalizes, and produces curated versions suitable for AI model training and inference.

Model developers manage the lifecycle of model-based artifacts. They experiment with different learning algorithms, train them on fused datasets that integrate multiple sensor inputs, and evaluate their predictive quality using performance metrics. Within the project, we observe that the most suitable forecasting model can vary depending on the prevailing weather system. This creates the challenge of either selecting a specific model for each system or running multiple models and applying ensemble strategies to produce the final predictions. In addition, model developers may execute training jobs on different computational resources, ranging from on-premise clusters to cloud infrastructures or supercomputers, which highlights the need for cross-platform support.

\begin{figure}[!ht]
\centering
\includegraphics[width=1.0\linewidth]{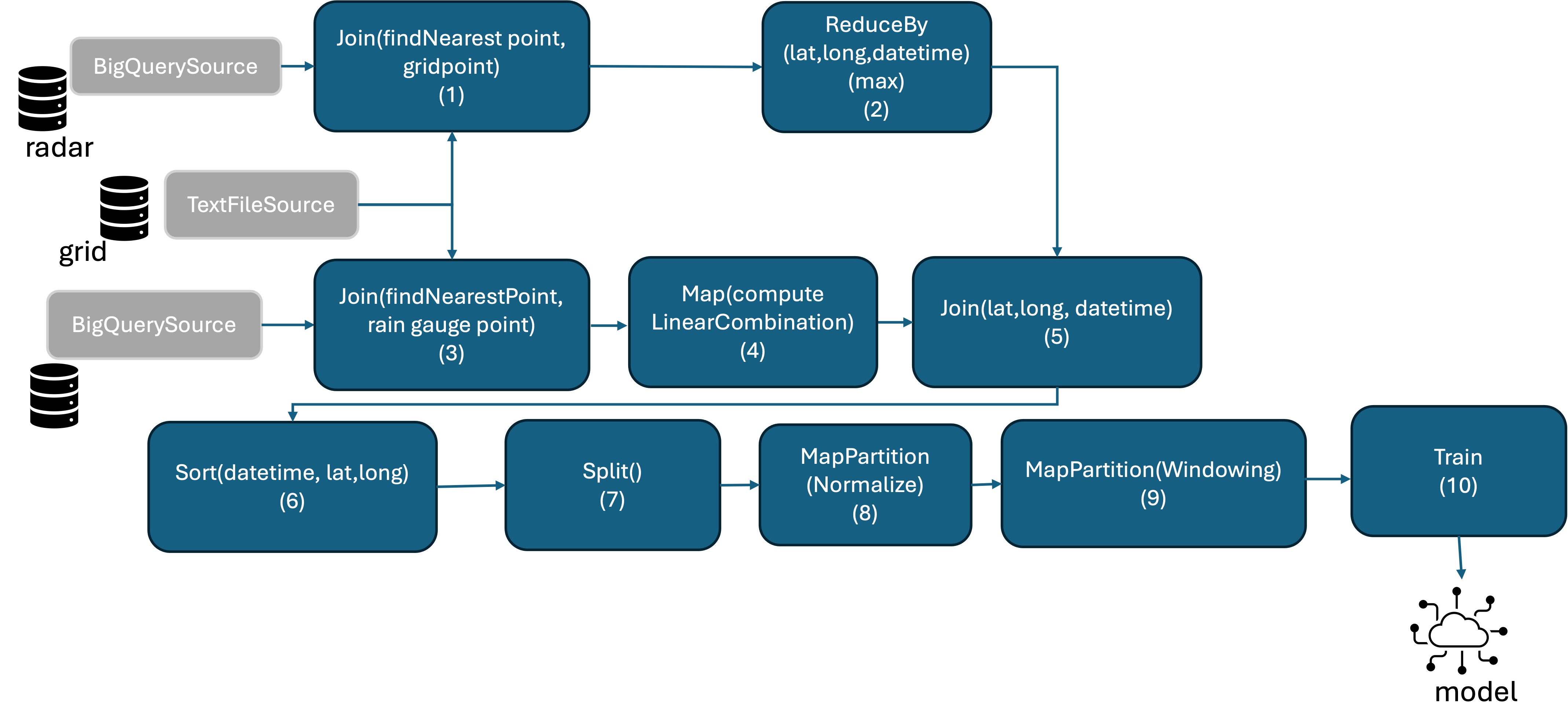}
\caption{\label{fig_buildModel} Dataflow for data preparation and model building}
\end{figure}

In Rionowcast, several learning algorithms have been trained under different data fusion configurations. For example, a ConvLSTM spatio-temporal model has been trained to forecast rainfall on a regular 2D grid with approximately 3 km spacing between cells, covering a 17 × 27 km rectangular region of Rio de Janeiro, with a temporal resolution of one hour. The model receives as input a multivariate time series \cite{ogasawara_event_2025}, combining precipitation measurements from rain gauges with radar reflectivity data. Figure \ref{fig_buildModel} illustrates the dataflow used to train the ConvLSTM model. In addition, another spatio-temporal model, STConvS2S \cite{castro_stconvs2s_2021}, has been trained using the same data preparation workflow (Figure \ref{fig_buildModel}), but adopting a 3D convolutional learning algorithm.

\begin{figure}[!ht]
\centering
\includegraphics[width=1.0\linewidth]{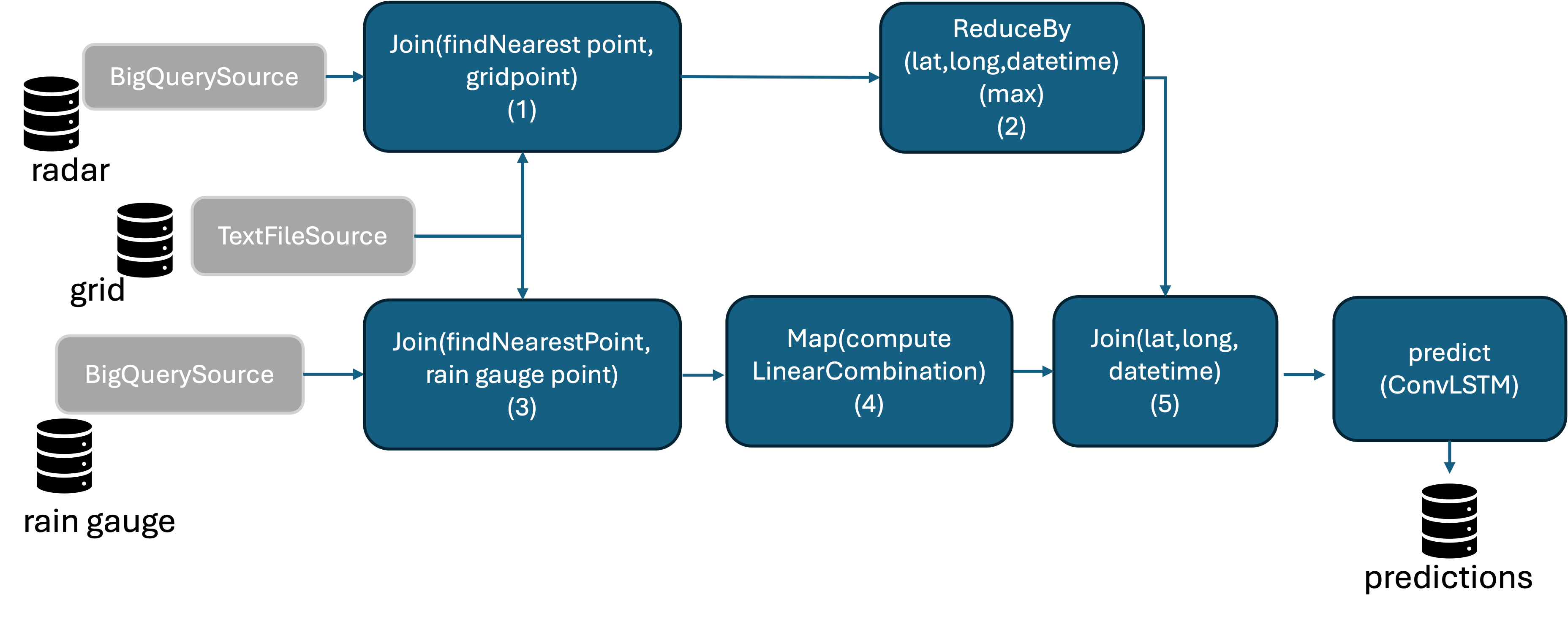}
\caption{\label{fig_inference} Data preparation for inference dataflow}
\end{figure}

Meteorologists monitor rainfall forecasts to issue extreme event alerts, using, for example, the inference dataflow in Figure \ref{fig_inference}. When a predicted rainfall rate exceeds 50 mm/hour, an alarm is triggered for a potential extreme rainfall event. To validate such alerts, meteorologists review streaming input data and cross-check them against other meteorological observations, such as online radar images. At this stage, they may also query datasets from additional weather sources or consult provenance information on the AI model construction to assess the reliability of the observed phenomenon.

In the Rionowcast project, Gypscie is used to provide data engineers and model developers with a unified view of all AI artifacts across their entire lifecycle, spanning heterogeneous platforms. It also offers meteorologists a high-level, web-based interface to this unified view, supporting both operational needs and scientific validation.

\section{Gypscie: System Overview}
\label{sec_architecture}
In this section, we give an overview of the Gypscie system, focusing on the user's perspective, the AI artifacts that are managed, the system architecture, data management and provenance management. The other major capabilities of Gypscie (model management, knowledge graph, and dataflow processing) are described in the next sections.

\begin{figure}[!ht]
\centering
\includegraphics[width=1.0\linewidth]{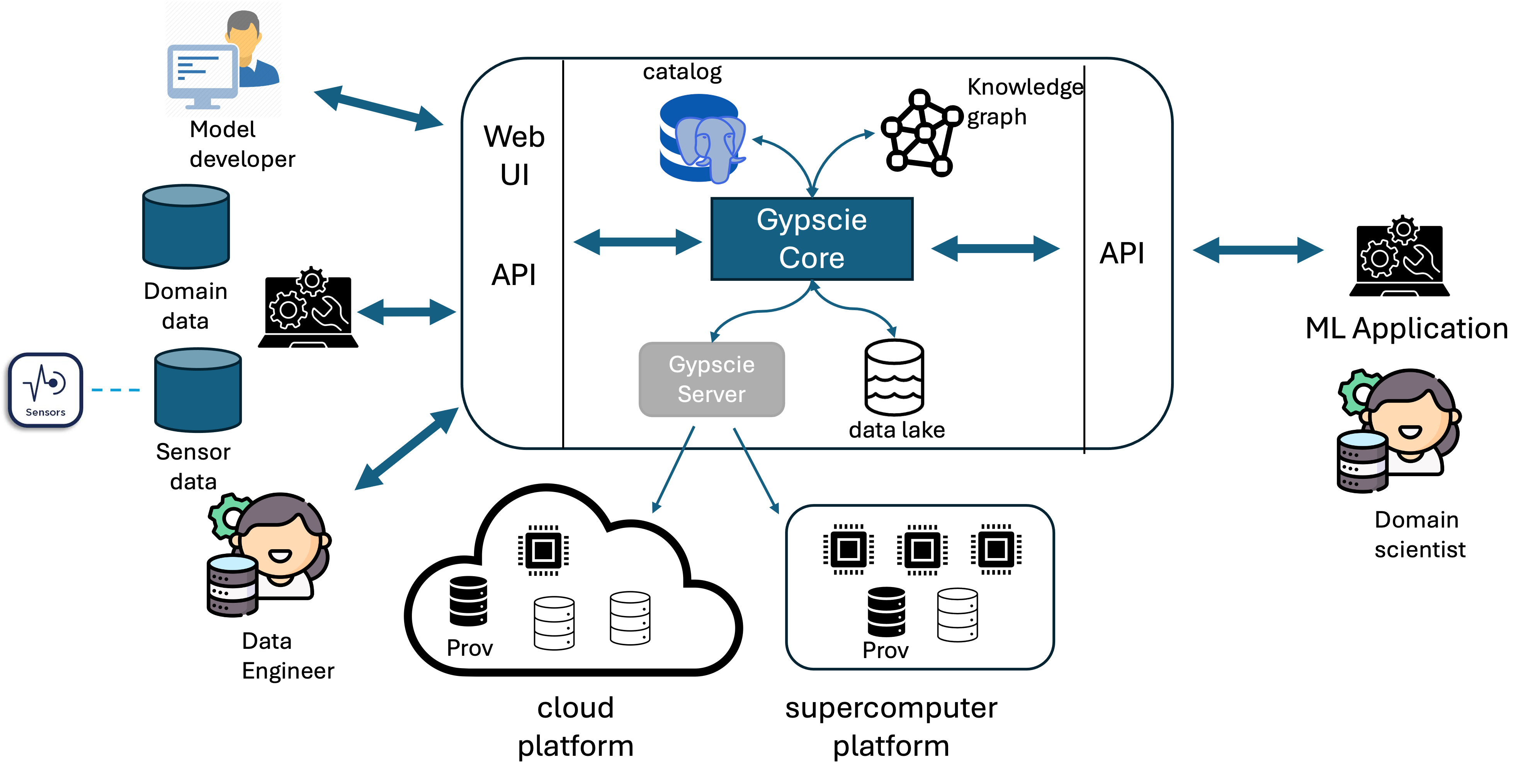}
\caption{\label{fig_gypscie_overview} Gypscie User’s Perspective}
\end{figure}

The requirements of potential Gypscie users have guided our design choices.
Figure \ref{fig_gypscie_overview} provides an overview of Gypscie, highlighting three main user roles: domain scientists (e.g., meteorologists) who analyze data using different models, model engineers who build and train models for the domain scientists, and data engineers who provide and maintain the datasets and dataflows for the model engineers. The figure also shows the main components of Gypscie from the user's perspective, namely Gypscie interfaces, the catalog, the knowledge graph, external data sources (domain and sensor data), AI applications, and access from the Gypscie core to supercomputers or cloud platforms. The three main user roles are as follows.

\begin{enumerate}
    \item 
\textbf{Model developers}. They carry out the model-oriented stage of the AI lifecycle, performing activities such as learner creation, model training, hyperparameter selection, model import, and model update. Model developers are also involved in combining trained models into ensembles and building new models from existing ones using transfer learning techniques.
    \item
\textbf{Data engineers}. They are in charge of the data-oriented stage of the AI lifecycle, involving activities such as data search, preprocessing, curation, and storage. They also create preprocessing dataflows. Data engineers manage data, define dataset schemas, associate them with a project and domain, and oversee storage, favoring data locality for efficient processing of dataflows. They also provide model developers with dataset metadata obtained from provenance records.
    \item
\textbf{Domain scientists}. They perform data analysis and prediction using an AI application's interface that relies on Gypscie's API to query data and use models. Through this interface, domain scientists can access predictions, domain knowledge, and provenance information to support their analysis and decision-making. 
\end{enumerate}
 
Gypscie also includes another user role: the AI engineer. This role monitors model performance in operation, looking for behavioral deviations that may indicate the need to retrain a model with new data. It combines responsibilities of the previously mentioned roles.

\subsection{AI Artifacts}
Gypscie manages the following main AI artifact types: datasets, models, functions, and dataflows. These are stored in the catalog as metadata. Datasets, models, and dataflows are qualified according to a given domain, which refers to a specific area of application. For instance, the datasets and models developed in the use case described in Section \ref{sec_example} are qualified under the Meteorology domain. Functions can be shared among domains. For example, a function that implements a specific data imputation algorithm \cite{miao_experimental_2023} may be useful for dataflows specified in various domains.

A Gypscie artifact has a unique Gypscie identifier (GID), which enables referencing it independently of any given context, such as the current host machine. A given artifact may have different versions, each identified by a version number as metadata. The choice of asserting that an artifact is a version of another is left to the user. Each artifact type is defined as follows.

\begin{itemize}
    \item 
\textbf{Dataset.}
A dataset is associated with a schema definition describing its attributes and attribute types. The dataset also includes its storage format (e.g., Parquet, H5, Zarr) and its registration date and time.
    \item 
\textbf{Model.}
The metadata include the type of task (e.g., classification, regression, clustering), the learning scope (supervised, unsupervised, semi-supervised, reinforcement learning), the creation date and time, and the dataset used for training.
    \item 
\textbf{Function.}    
A function implements one of the following behaviors: data transformation, model construction, or dataset input or output.  
Two types are distinguished for input and output: source and sink. A source function reads a stored dataset and creates a version of it in memory, whereas a sink function reads an in-memory dataset and writes it as a stored dataset. Data transformation functions read one or more in-memory datasets and write a single output dataset. A model construction function is referred to as a learner, which reads an in-memory training dataset and produces a stored trained model.  
    \item 
\textbf{Dataflow.}
Data engineers or model developers implement data processing, training, and inference tasks as dataflows. A dataflow specifies a directed acyclic graph (DAG) of data dependencies between its operator nodes (see Section \ref{sec_dataflow}).

\end{itemize}

\subsection{Interfaces}
\label{sec_gypscie_interfaces}
Gypscie offers two interfaces: a web interface and an API. The web interface provides simple access to all Gypscie services, while the API enables the integration of Gypscie services within more complex applications. For instance, in a meteorology application, front-end pipelines that consume data from different streaming sources can use the Gypscie dataset registration API to register data windows. The meteorology panel used by meteorologists can then consume the predictions produced by AI models in Gypscie, using the prediction query API. 

For more complex services involving multiple data transformations and AI model-related actions, Gypscie offers an interface to create and manage dataflows and functions (see Section \ref{sec_dataflow}). Gypscie uses an internal interface to submit dataflows for execution and to retrieve metrics, results, and artifacts.  
The core module sends a bundle to a target AI platform (e.g., an execution queue in the Santos Dumont supercomputer) containing a JSON materialization of the dataflow specification, the GID of the input datasets, the code for the dataflow functions, and a container with the necessary libraries.

\subsection{System Architecture}

Gypscie leverages recent advancements in web-oriented architectures. We adopt the following design principles:

\begin{itemize}
\item Ease of use through web interfaces;
\item Service-based interoperability via well-defined APIs;
\item Heterogeneity management through support for multiple AI libraries, learner types, data formats, and data representations;
\item Cross-platform support enabling execution across clusters, supercomputers, and cloud infrastructures;
\item Provenance management to support explainability and reproducibility of tasks;
\item Integrated data view provided by a knowledge graph with a rule-based query capability.
\end{itemize}

\begin{figure}[!ht]
\centering
\includegraphics[width=1.0\linewidth]{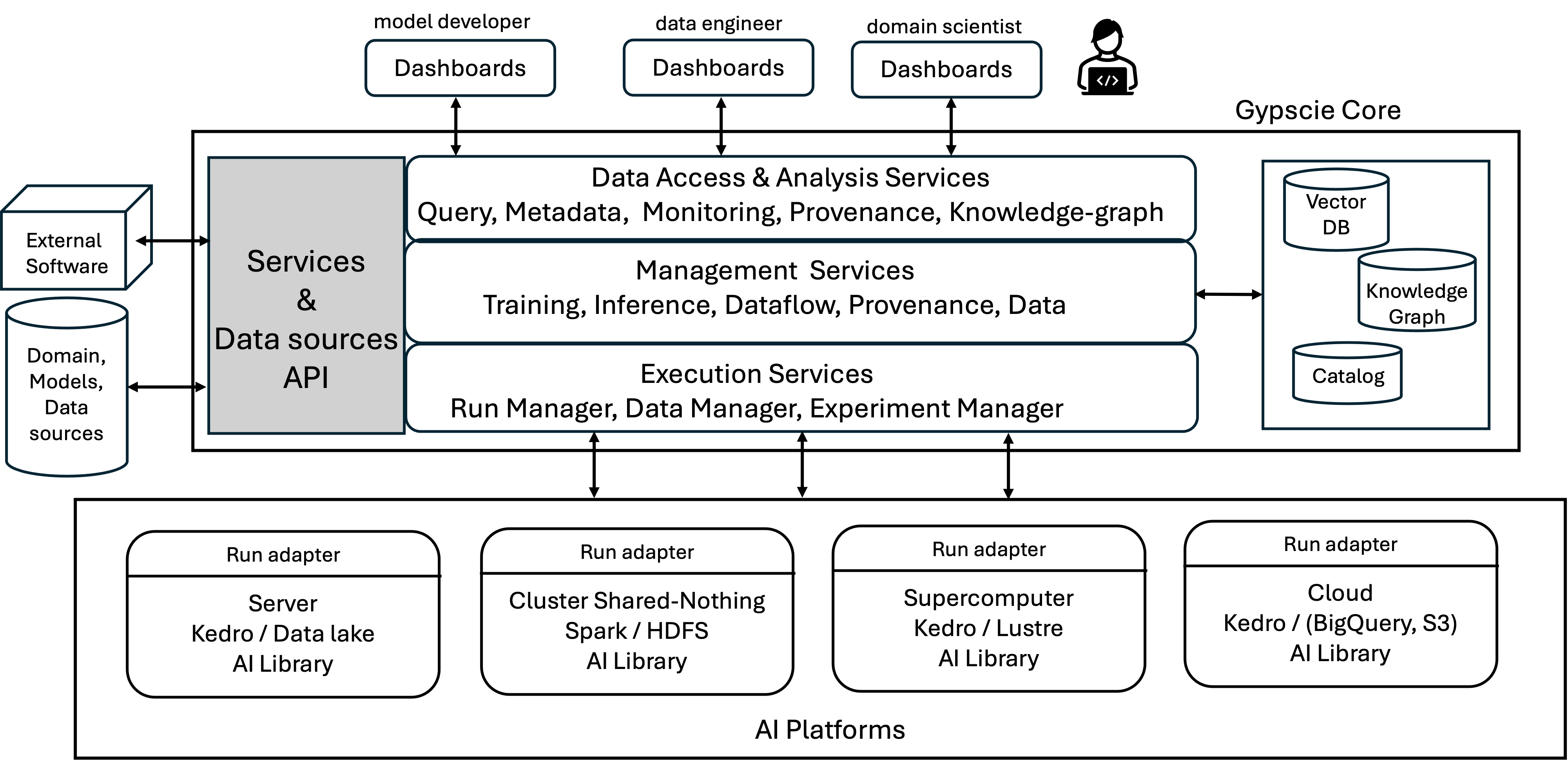}
\caption{\label{fig_gypscie-archi} Gypscie Architecture}
\end{figure}

The architecture is divided into two main components, as shown in Figure \ref{fig_gypscie-archi}: the core module and the AI platforms module. The core module orchestrates the execution of service requests, including preparation, scheduling, and result retrieval.

An AI platform provides the infrastructure that executes requests, such as a server, cluster, supercomputer, or cloud. It consists of hardware for computation and artifact storage, along with the libraries required to run AI tasks. The libraries needed for dataflow functions are provided through containers. Each AI platform is registered in the Gypscie catalog, which makes it available for job scheduling. The AI platform communicates with the core module through an adapter API that unifies the interface between the Gypscie core and the various platforms, ensuring the execution of service requests.

For external communication, the core component interacts with Gypscie applications through its service API. This API allows external applications to register artifacts and service requests in Gypscie and to retrieve the results of executed services. 

Datasets produced by external sources are registered in Gypscie to obtain a GID and to make them eligible for access by service requests. If a dataset is already stored on a Gypscie AI platform, registration only requires generating a new GID and recording its provenance information. For example, a dataset produced by an application running on a supercomputer and stored in its file system would only need GID assignment and provenance registration. Conversely, if the dataset is stored on a device that does not belong to a registered AI platform, it must be uploaded to Gypscie during the registration process. A similar procedure applies to models built on an AI platform not registered in Gypscie. In this case, the model is registered using Gypscie's import service, which assigns the model a GID and follows the same steps as for datasets regarding physical storage and provenance information.

The main services offered by Gypscie are as follows:
\begin{itemize}
    \item Artifact management for registration, metadata and provenance management, artifact updates, and exclusion operations;
    \item Model training to build a model with a learner, an input dataset, a set of hyperparameter values, and an AI platform;
    \item Model inference using a trained model, an input dataset, parameter values, and an AI platform to compute predictions; 
    \item Model import to register a model built outside Gypscie;
    \item Query to search the knowledge graph for artifact metadata, domain data, predictions, and provenance information about artifacts and service execution;
    \item Data management to organize and store datasets;
    \item Dataflow definition, see Section \ref{sec_dataflow}, including a list of input dataset GIDs, the names of output datasets, and an AI platform; 
    \item Automatic scheduling and optimization of dataflows to analyze execution requests and to decide on the AI platform where they should be run;
    \item Provenance management to capture, store, and query provenance information about artifacts.
\end{itemize}

\subsection{Data Management}
\label{sec_datasetmanagement}
Gypscie manages datasets throughout the lifecycle of AI artifacts. A dataset is a collection of data elements. For instance, a rainfall dataset consists of a set of rainfall recordings. Each element may be of any data type, simple or structured. Thus, a dataset of precipitation time series may contain a list of floating-point values or a set of tuples of measurements including latitude, longitude, precipitation, humidity, and pressure.

When a dataset is registered, it receives a GID along with a schema describing its attributes. Currently, Gypscie supports the following data storage formats: Parquet, CSV, NetCDF, H5, Zarr, and HDFS. Other formats can be integrated as needed. The files themselves are stored in file systems such as Apache HDFS, Lustre, or Amazon S3, depending on the AI platform.

Datasets are stored either in data lakes or file systems, depending on the resources available within a given AI platform. To organize storage, Gypscie structures the data space into three buckets: landing, staging, and curated.
\begin{itemize}
    \item The landing bucket contains datasets extracted from their original data sources but not yet processed. For example, a dataset window captured from a meteorological radar is first copied into the landing bucket.
    \item The staging bucket stores preprocessed datasets. At this stage, errors may have been corrected and a standard dataset format may have been applied.
    \item The curated bucket holds datasets that have undergone all necessary preprocessing and are ready to be used in dataflows. For time-series datasets, the curated bucket is further divided into real-time and historical subfolders. Real-time datasets are used directly as inputs to real-time inference dataflows, whereas historical datasets are preserved as training data.
\end{itemize}

Gypscie prioritizes data locality, storing datasets within the AI platform's native file system where they were created. It also provides a default AI platform to store datasets originating from external data sources registered in Gypscie.

The placement of a dataset significantly affects job execution performance. When a dataflow execution request is submitted to Gypscie, it specifies the input datasets required for processing the corresponding job. The job scheduler then selects the AI platform on which to execute the job and allocates the corresponding datasets. In some cases, datasets must be copied from their storage platform to the platform where the job is scheduled to run, which introduces performance overhead. To mitigate this, the Data Management service periodically analyzes dataset access patterns to predict the most likely platform from which a dataset will be accessed. Based on this analysis, it may proactively replicate the dataset to the target platform. Gypscie also manages dataset versions across platforms, removing redundant copies when they are no longer accessed locally.

\subsection{Provenance Management}

Gypscie implements services to capture, store, and query provenance information about its artifacts and their roles in dataflow execution. Provenance information encompasses both prospective provenance, structured in the Gypscie catalog, and retrospective provenance, which is generated during dataflow execution. In prospective provenance, metadata about dataflows, functions, learners, datasets, and models are recorded. In retrospective provenance, three types of records are distinguished, artifact generation, training metrics, and user-defined records \cite{bastos_twinscie-prov_2025}. Provenance metadata enable the reproducibility of experiments, support dataflow optimization, and contribute to prediction explainability and automatic model selection.

Datasets serve as both inputs and outputs of dataflow operations. The provenance of every dataset registered in Gypscie is recorded in the system's provenance database. For datasets originating from external systems, their source description is stored as provenance information. For datasets produced by Gypscie dataflow operations, the provenance includes references to the dataflow, the operation, the function, and the input datasets used in their creation.

During artifact generation, Gypscie leverages MLflow to retrieve results and associate them with the functions responsible for their production. Training metrics are automatically captured via Autolog, an MLflow feature, with values recorded at each epoch and sent to the Gypscie Provenance Manager to update the provenance database.

User-defined records, such as variable statistics (e.g., mean and standard deviation), can be added through configurable hooks in dataflow nodes, as described in Section \ref{sec_dataflow}. These records are exported in JSON, stored in the MLflow server, and later integrated into the Gypscie catalog. Provenance metadata are stored in the Gypscie catalog, which has been extended to support provenance records. Datasets stored in AI platforms are also referenced by retrospective provenance metadata as part of the provenance artifacts.

Finally, a provenance graph, implemented according to the W3C PROV standard \cite{moreau_prov-overview_2013}, can be generated upon request by a domain scientist or through the Gypscie API.

\section{Model Management}
\label{sec_Modelmanagement}

Model management plays a vital role throughout the AI lifecycle. It supports activities in both the model-oriented and operations stages and indirectly influences the data-oriented stage \cite{schlegel_management_2023}. Building a model is an iterative process \cite{idowu_asset_2023}, involving tasks such as preparing input datasets, configuring hyperparameters, selecting learning algorithms, defining the scope of the learning task (e.g., supervised or unsupervised), choosing the AI training tools, and scheduling dataflows on the AI platform. Not every activity must be repeated in each iteration. For instance, a model engineer may reuse existing artifacts while only adjusting hyperparameters for a new training run.

In this section, we describe how Gypscie supports model management activities, including learner management, versioning, training, inference, and model selection.

\subsection{Learners}
In Gypscie, the learner is a central artifact in model training. Implemented in Python, it encapsulates key design decisions such as the choice of learning algorithm, model architecture, expected input data format, criteria for splitting data into training, validation, and test sets, and the metrics used to evaluate model quality. The learner serves as both a reusable component and the primary unit for attaching provenance information.

\subsection{Model Versions}
Managing AI models involves handling various model versions and copies. Given a model, the model manager creates a new model version when one of the following conditions occurs:
\begin{itemize}
    \item Hyperparameters are adjusted (e.g., learning rate, batch size, or number of trees).
    \item The training dataset is updated (with more data, cleaner data, or rebalanced data).
    \item The training process changes (e.g., number of epochs or optimization settings), while the underlying architecture or algorithm remains the same.
    \item Minor code refactors are performed without altering the model's structure or learning paradigm.
\end{itemize}

Managing different model versions prevents unwanted impacts on applications, such as recompilation, when a new version is deployed. The model developer may also decide to discard a specific version, which has no effect on applications. The management of model copies is distinct. A model copy corresponds to the same model artifact stored on more than one AI platform. The model manager maintains multiple copies to enhance data locality during inference dataflow scheduling. Conversely, a copy is discarded when it is no longer useful for scheduled dataflows.

In contrast to a model version, a new model is built when one of the following conditions is observed:
\begin{itemize}
    \item The learning algorithm changes (e.g., switching from random forest to extreme gradient boosting, or from long short-term memory (LSTM) to transformer).
    \item The model architecture changes in a way that alters its inductive bias (e.g., modifying the number or type of layers in a neural network or adding attention mechanisms).
    \item The problem definition changes (e.g., from regression to classification, with different target variables or input features).
    \item The task or domain changes (e.g., from rainfall forecasting to traffic forecasting).
\end{itemize}

These activities, including producing a new model version or building a new model, are supported by Gypscie's training service.

\label{sec_processing_training}
A model training request can be issued by an external Gypscie component through its service API or directly by a domain scientist via the web user interface. In both cases, the request is forwarded to the model manager, which then communicates with the Gypscie core run manager, the component responsible for submitting jobs to the selected AI platform.

Once the training job is completed, the model manager uses the job ID and training logs to update the Gypscie catalog. The job ID is also passed to the provenance manager to retrieve provenance metadata, including model metrics. The resulting model artifact is stored in the local storage of the AI platform used for training, and its location is recorded in the logs and updated in the Gypscie catalog. Once the model artifact is generated, it becomes available for execution in inference tasks.

\subsection{Model Inference}
\label{sec_model_inference}
Model inference involves selecting a model artifact, an input dataset formatted according to the model's expected input, and an AI platform. The model manager handles the inference request by interfacing with the run manager to execute the inference dataflow and by updating the catalog with metadata about the resulting prediction file.

\subsection{Model Selection} 
\label{sec_model_selection}
Model selection enables users to search for model artifacts of interest for inference. Searches can be performed based on various criteria, such as scientific domain or subdomain, metadata, format, tools, and keywords. This capability relies on the artifact catalog.

A key functionality of the Gypscie model manager is the automatic selection of models for a given prediction task. Automatic model selection addresses the problem of identifying the optimal model from a set of existing models designed for the same task. These models may represent different versions of the same model or entirely different models. Gypscie provides extensible system-level support for model selection within a broader cross-platform AI artifact management architecture, allowing different selection strategies to be integrated with artifact metadata and provenance management. To address this task, the model manager uses the DJEnsemble selection algorithm \cite{pereira_djensemble_2021}. The intuition behind DJEnsemble is to rank models according to the similarity between the data used during their training and the input data provided for inference. This approach leverages the provenance information maintained by Gypscie for all trained models. In addition, DJEnsemble evaluates each model's generalization ability through a meta-learning function. In the spatio-temporal context of Rionowcast predictions, the model selection procedure is combined with a spatial allocation strategy, following the approach described in DJEnsemble.

\section{Knowledge Graph}
\label{sec_knowledge-graph}

In this section, we provide a detailed description of our knowledge graph approach.

A knowledge graph \cite{hogan_knowledge_2021} provides a flexible and semantically rich way to integrate heterogeneous data sources by representing entities and their relationships in a unified graph structure. Unlike rigid schema-based integration approaches, knowledge graphs can naturally accommodate diverse formats, evolving schemas, and complex interconnections. By embedding semantic meaning through ontologies and linked vocabularies, they enable more accurate data alignment, support reasoning across data sources, and improve discoverability. This makes them particularly effective for integrating siloed or rapidly changing data in complex domains such as science, where context and relationships are as important as the data itself.

In Gypscie, we adopt a knowledge graph approach to provide an integrated view of the data and metadata needed by AI applications. The data considered for integration include historical training data, static domain data, online streaming data, observations, predictions computed by AI models, and provenance data. The knowledge graph exposes to AI applications the data and relationships that are otherwise hidden within the different data sources managed by Gypscie. It includes managed artifacts such as datasets, models, learners, functions, and dataflows, as well as execution and provenance entities such as model runs, transformation runs, and dataflow runs. The graph captures relations such as artifact usage, input and output dependencies, domain qualification, and provenance links between produced artifacts and the operations that generated them. Rule-based reasoning extends this explicit graph with derived relationships, enabling the system to expose not only stored metadata but also inferred connections between artifacts and executions.

\begin{figure}
\centering
\includegraphics[scale=0.35]{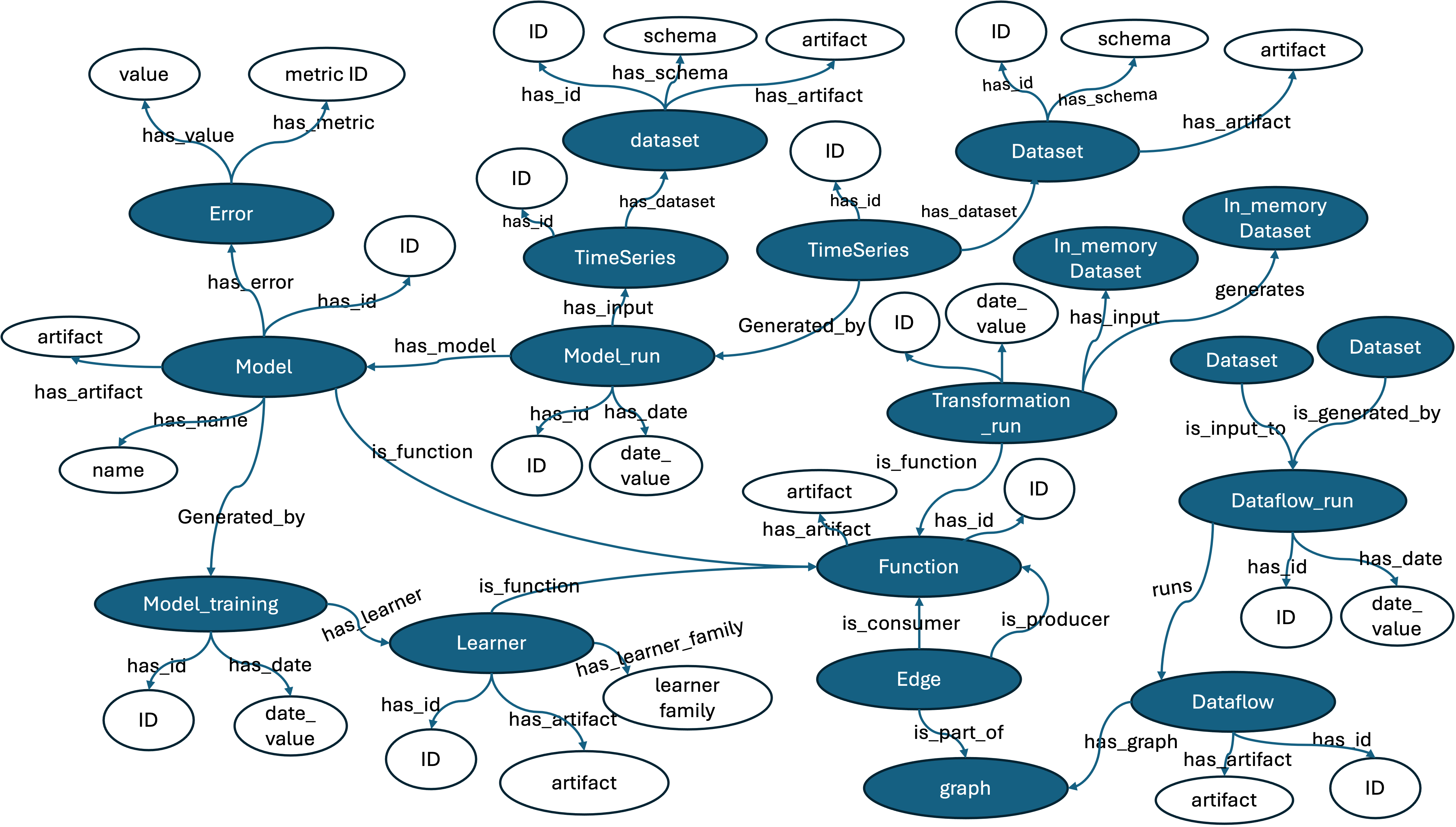}
\caption{\label{fig_Knowledge_graph} Gypscie Knowledge Graph}
\end{figure}

Figure \ref{fig_Knowledge_graph} shows a graphical representation of a fragment of the Gypscie knowledge graph for the meteorology use case described in Section \ref{sec_example}. We can identify nodes representing managed artifacts such as models, learners, datasets, and dataflows. Provenance information includes model training records and errors, model runs representing inference executions, transformation runs, and dataflow runs. The graph also includes datasets containing input time series for model prediction, as well as time series generated by the inference process.

A knowledge graph can be queried using graph-based languages such as SPARQL (for RDF data) or Cypher (for graph databases). In Gypscie, we also need to manipulate complex domain knowledge, which is ubiquitous in scientific areas such as physics, medicine, and meteorology. Therefore, the query language must also enable the expression of domain knowledge to extend the information obtained from the available data. This knowledge semantics can be naturally captured by a rule-based language grounded in first-order logic, such as Datalog.

In Gypscie, we adopt the approach used in InteGraal \cite{baget_integraal_2023}, a tool for integrating heterogeneous data sources using Datalog conjunctive queries. The combination of a knowledge graph with a rule-based language provides several advantages \cite{moraes_gypscie-kg_2025}: a unified representation for expressing queries and dataflows, the ability to define domain rules as extensions of domain knowledge, and cost-based optimization of queries and dataflows. In this setting, the explicit graph structure provides the base relations, whereas Datalog rules define derived predicates that capture higher-level semantics needed by AI applications.

The code snippet in Listing \ref{lst:knowledge_graph} illustrates rules specified in the InteGraal knowledge base integration system \cite{baget_integraal_2023}. Rules in InteGraal extend the Global as View integration approach with existential semantics defined over the integrated view. The existential semantics enables the definition of concepts that do not exist in the integrated view. Consequently, a domain scientist can write queries on the knowledge base that reason over these defined rules. In Listing \ref{lst:knowledge_graph}, the \textit{is\_activity} predicate generalizes the different processes executed in Gypscie. The transitivity rule navigates through input and output relationships to produce the transitive closure of dataset transformations. 

\begin{lstlisting}[basicstyle=\ttfamily\scriptsize, caption={Example of rules in InteGraal}, 
        label={lst:knowledge_graph},
        showtabs=false,
        showspaces=false]
is_activity(IDMR):- model_run(IDMR).
is_activity(IDMT):- model_training(IDMT).
is_activity(IDTR):- trans_run(IDTR).

% Transitivity 
transformation(IDDSI,Name,IDDSO) :- has_input(IDTR, IDDSI),
uses(IDTR,IDTF), has_name(IDTF,Name),has_output(IDDSO,IDTR),
dataSet(IDDSO).
transformation(X,ig:concat(W1,ig:concat(" + ",W2)), Z) :- 
transformation(X,W1 ,Y), transformation(Y,W2, Z).
\end{lstlisting}

This formalization is particularly useful in Gypscie because it connects lifecycle management with semantic querying. For example, the same integrated view can be used to retrieve the datasets and models involved in a prediction, to trace transformation chains through provenance relations, or to support optimization decisions based on artifact dependencies.

\section{Dataflow Management}
\label{sec_dataflow}

Dataflows are high-level representations of data pipelines that automatically move and transform data from their sources to where they are needed. They are typically used for data collection, data cleaning and preprocessing, feature engineering, data validation, and data transformation. They are critical because raw data from real-world systems are rarely clean or ready to be used directly by an AI algorithm. A dataflow ensures that data flows reliably and consistently through every preparation stage. We distinguish three levels of dataflow representation: abstract dataflows, concrete dataflows, and data pipelines.
\begin{itemize}
    \item An abstract dataflow defines a directed acyclic graph (DAG) of dependencies among operations applied to dataset placeholders.
    \item A concrete dataflow instantiates an abstract dataflow by binding it to specific datasets as inputs and outputs and by providing the parameter values required for execution.
    \item A data pipeline refers to a concrete dataflow in execution as a job running on a given AI platform.
\end{itemize}
This section describes our high-level dataflow language and our approach to dataflow processing and optimization.

\subsection{Dataflow Language}
Our dataflow language represents data transformations as a DAG. It is designed to achieve the following goals: enable automatic optimization, provide explainability for dataflow outcomes, express services spanning the complete AI lifecycle, and execute dataflows on different AI platforms.

To enable automatic optimization, we consider dataflow nodes associated with operators that have well-defined data transformation semantics, such as filter, join, or train. The metadata of these nodes are further enriched during dataflow execution with provenance information whose logs describe the execution behavior of each operator, thereby supporting cost-based optimization. The provenance data captured during operator execution also provide information about dataset transformations, enabling explainability of executed processes. 

Representing a dataflow as a DAG enables diverse compositions of operations. The behavior of each operation is defined by an associated user-defined Python function that specifies the expected behavior of the operator within the dataflow. This compositional flexibility supports the definition of diverse services across the AI lifecycle.

The transformation from an abstract to a concrete representation for dataflow execution adapts the operator code to the language used by the target AI platform on which the dataflow will run. Examples of such execution frameworks include Apache Spark, PyTorch, TensorFlow, and Velox.

\begin{figure}[!ht]
\centering
\includegraphics[scale=0.2]{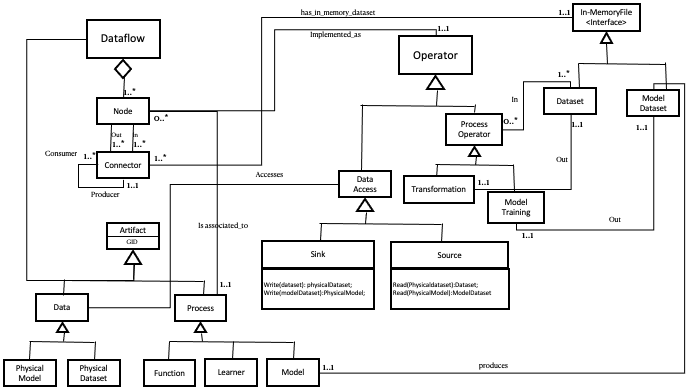}
\caption{\label{fig_Abstract} Dataflow language data model}
\end{figure}

Figure \ref{fig_Abstract} shows a UML representation of the dataflow language data model. An artifact is a generic concept that represents data, processes, and dataflows. Each artifact object is identified by a unique GID, as described in Section \ref{sec_architecture}. 

Dataflow nodes are associated with operators that specify their semantics in one of three categories: data access, transformation, and model training. Data access operators are of the form $DA = \{sink, source\}$. A $source$ operator reads an input file (i.e., a physical dataset) and produces a copy of its content in memory (i.e., an in-memory dataset). Conversely, a $sink$ operator stores an in-memory dataset in persistent storage.

A transformation is an operation that processes input in-memory datasets and produces an output in-memory dataset. The execution behavior of a transformation operator includes its input–output dataset cardinality, denoting the number of datasets that are input to and output by the operator. It also specifies whether it processes each dataset element one at a time or assumes a set-at-a-time behavior. For example, the $Map(dataset\ in, dataset\ out, function\ f)$ operator implements element-at-a-time behavior, whereas the $GroupBy(dataset\ in,\\ dataset\ out, attribute\_list\ param)$ operator implements set-at-a-time behavior.  

The \textit{Process} type in Figure \ref{fig_Abstract} generalizes the function object that defines the processing behavior of operators. A \textit{Transformation Process} operator is associated with a \emph{Function} process implemented in a programming language such as Python. The semantics of operators regarding set-at-a-time versus one-element-at-a-time processing restrict the type of functions compatible with them. The \textit{Model Training} operator is of type \textit{Process Operator}. Examples of Model Training operators are $MT=\{train, fit\}$. The \textit{Process} used in model training is of type \emph{Learner}. The latter refers to code in a programming language (e.g., Python) that implements a learning strategy and produces an output model artifact. Finally, a \textit{Model Process} operator is associated with \textit{Transformation} operators $Mo=\{Predict, Infer, \ldots\}$, which receive input data and compute inference results as output.

Listing \ref{lst_DSL_abstract} shows an abstract dataflow for the Rionowcast inference use case, corresponding to the dataflow definition in Figure \ref{fig_inference}, with a few simplifications. The function implementations are not included. Such a definition is used in several concrete dataflows. When specifying a concrete dataflow, input and output datasets are associated with the corresponding $sink$ and $source$ operators, and function parameters are assigned values. A concrete dataflow is then scheduled to run as a pipeline on a specific AI platform. 

\begin{lstlisting}[language=Python, basicstyle=\ttfamily\scriptsize, caption={Abstract dataflow
for the Rionowcast inference use case}, 
        label={lst_DSL_abstract}]
graph_data = [
    {
        "GID":"hash_id",
        "description":"Rionowcast_inference"
    },
    {
        "operator":"source",
        "function_alias":"rionowcast_inference_load_data",
        "input":[
            "radar_data_path",
            "rain_gauge_data_path",
            "grid_data_path"
        ],
        "output":[
            "df_radar", 
            "df_rain_gauge",
            "df_grid"
        ],
    },
    {
        "operator":"source",
        "function_alias":"rionowcast_inference_load_model",
        "input":["model_path"],
        "output":["model"],
    },
    {
        "operator":"Join",
        "function_alias":"rionowcast_inference_preprocessing",
        "input":[
            "df_radar", 
            "df_rain_gauge",
            "df_grid"
        ],
        "output":["X"],
    },
    {
        "operator":"predict",
        "function_alias":"rionowcast_inference_predict",
        "input":[
            "X",
            "model",
        ],
        "output":["output"],
    },
    {
        "operator":"sink",
        "function_alias": "rionowcast_inference_save",
        "input":[
            "output",
            "output_path"
        ],
        "output":None,
    }
]
\end{lstlisting}

Since an operator may consume more than one dataset, we need to disambiguate the pair that establishes the producer–consumer relationship between two nodes. This relationship is represented by the pairing of dataset connectors, each corresponding to a specific input or output dataset.

We consider two types of functions, $\theta$ and $\lambda$. $\theta$ functions are transformation functions. They may receive a single tuple as input and produce a single tuple as output. Transformation functions may also behave as aggregate functions, receiving a set of tuples as input and producing a single output tuple. $\lambda$ functions implement model training, receiving a set of tuples as input and producing a model and an output tuple with metric values.

\begin{figure}[!ht]
\centering
\includegraphics[scale=0.2]{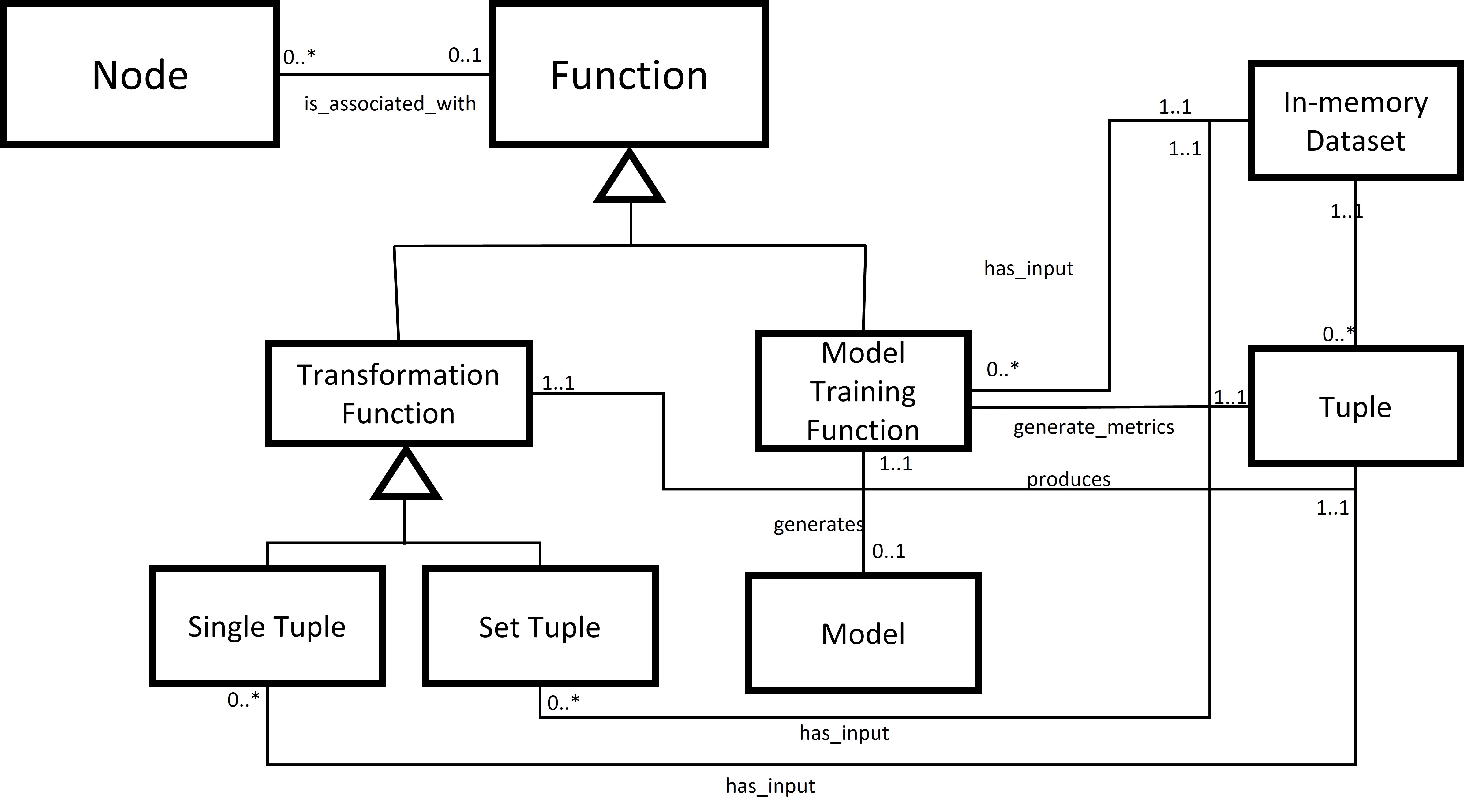}
\caption{\label{fig:function} Function data model}
\end{figure}

Figure \ref{fig:function} complements Figure \ref{fig_Abstract} by focusing on the definition of two types of transformation functions that operate on either single input tuples or sets of input tuples. The figure also shows the \textit{Model Training} function (i.e., a learner), which receives an in-memory dataset as input and generates a model artifact and a tuple of training metrics. Finally, a concrete dataflow associates the GIDs of each input dataset with the input connectors of the source operators.

\subsection{Dataflow Processing}
\label{sec:dataflow_processing}
Dataflow processing in Gypscie includes dataflow rewriting, scheduling, and execution.
At a high level, given an abstract dataflow $D$, a set of fragments $F$ derived from $D$, and a set of available AI platforms $P$, the scheduling problem can be viewed as finding an assignment $A: F \rightarrow P$ that minimizes the total cost of executing and transferring dataflow fragments while satisfying platform constraints such as data locality, execution compatibility, and GPU availability. This objective can be expressed as follows:
\[
\mathrm{Cost}(D,A)=\sum_{f \in F}\mathrm{execution\_cost}(f,A(f))+\sum_{(f_i,f_j)}\mathrm{transfer\_cost}(f_i,f_j,A)
\]
where the first term captures fragment execution cost on the assigned platform and the second term captures the cost of moving data between dependent fragments assigned to different platforms. In Gypscie, this objective is approached heuristically, with provenance information providing the basis for cost and selectivity estimation.
In practice, Gypscie approaches this objective in two main steps: dataflow rewriting, which improves the execution plan using operator semantics, and platform-aware scheduling, which assigns the resulting fragments to target AI platforms.

Dataflow rewriting involves ordering operators in the dataflow using a strategy inspired by \cite{ogasawara_algebraic_2011}. We leverage the semantics introduced by operators at each node to evaluate rewriting rules that modify their positions in the DAG. For example, a filter operator can be pushed as close as possible to the source operators that produce the data it filters. We adopt the predicate migration technique, applied to expensive predicates in SQL queries \cite{hellerstein_predicate_1993}, to generalize the movement of operator nodes. This technique computes a rank for operations based on the ratio between operator selectivity and cost. Higher-ranked operators are placed as early as possible in the graph path. We use provenance records as the basis for selectivity and cost estimation.

For performance or data privacy reasons, a dataflow in Gypscie may have to be executed on different AI platforms. For example, data transformations may occur in a cluster where the data are stored, while model training may take place on a supercomputer. Thus, we introduce the concept of a \textit{dataflow fragment} as a unit of scheduling for a registered AI platform. We define a dataflow fragment as a connected subgraph of a dataflow that starts at a \textit{source} operator and ends at a \textit{sink} operator, along with their corresponding input and output files.

\begin{figure}[!ht]
\centering
\includegraphics[scale=0.08]{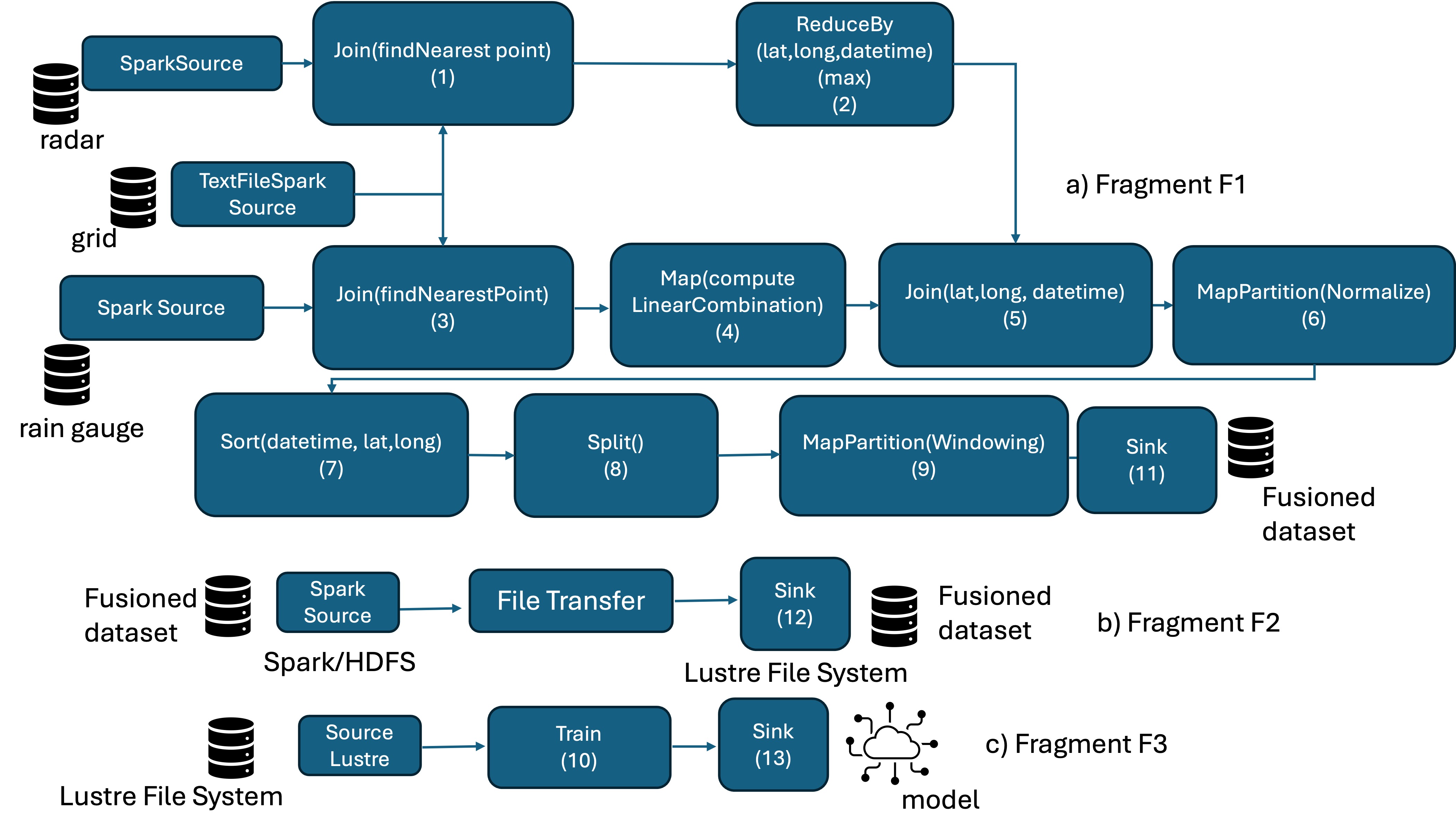}
\caption{Dataflow split into fragments}
\label{fig:fragments}
\end{figure}

Dataflow scheduling involves splitting the dataflow into fragments, introducing data transfer fragments, and assigning each fragment to an AI platform. Consider, for instance, the dataflow shown in Figure \ref{fig_buildModel}. Suppose the input datasets, radar, rain gauge, and grid files, are stored in a Spark CPU-based cluster. The preprocessing operations to build a training dataset can exploit data locality by executing on the Spark cluster. Because of the large size of the fused dataset resulting from preprocessing the three files, the training step can be scheduled to run on a supercomputer AI platform equipped with multiple GPUs. The execution plan would therefore split the dataflow into three fragments, as shown in Figure \ref{fig:fragments}. Fragment (a) includes all operators from reading the input files to executing preprocessing steps and storing the fused dataset. Fragment (b) implements the transfer of the fused dataset to the supercomputer file system. Finally, fragment (c) contains the training operation.   

This complex activity is usually performed manually by the dataflow developer, which is error-prone and suboptimal. In Gypscie, we automate this process using a scheduling approach that prioritizes data locality for data preprocessing and model inference and GPU availability for model training. The computation of dataflow fragments proceeds as follows. It starts by analyzing the dataflow graph and marking nodes with operator types \emph{source} and \emph{train}. For the former, we identify the current storage location for each input dataset. We then traverse the graph from the dataset \emph{source} operation forward, following the direction of the graph edges, until a data transfer is required or a \emph{sink} operation is reached. The fragment of the graph produced up to that point is annotated with the corresponding dataset location and AI platform. Operations requiring data transfer are either binary operators, which associate fragments from different AI platforms, or the \emph{train} operator. In this process, the operator semantics introduced earlier, together with the distinction between transformation functions $\theta$ and training functions $\lambda$, provide the formal basis for reasoning about fragment boundaries, execution requirements, and platform suitability.

If a binary operator is found, three options are possible: execute the binary operator at the site of one of the input fragments, at the site of the next binary operator, or at the train operator site. This choice considers dataset transfer costs, using provenance information for cost-based decisions. The \emph{train} operator is scheduled on a GPU-equipped AI platform and annotated with the number of GPUs to be used during model training. Suppose the input fragment assigned to the train operator is already located on an AI platform satisfying the GPU requirements. In that case, the input fragment is extended to include the train operator. Otherwise, the latter is assigned to the AI platform with GPU units that best match the requirements. In this case, a dataset transfer fragment is added to the schedule, which performs the data copy from the input fragment site to the training site. Data transfer operations use the RDMA protocol \cite{sur_rdma_2006} to copy datasets between AI platforms without staging large datasets to disk. Thus, transfer operation nodes are split between the input fragment and the fragment at the transfer destination site.

Finally, the output of dataflow optimization is the dataflow execution graph split into fragments and annotated with the assigned AI platforms. Next, the AI platform dataflow adaptation rewrites the graph into a specification supported by the target AI platform. We currently support execution in Apache Spark, Kedro, Parsl, and the AI engines TensorFlow, PyTorch, and Scikit-learn. The mapping procedure is inspired by the Apache Wayang project \cite{beedkar_apache_2025}. Finally, the mapped code is scheduled for execution as a set of data pipelines to be run as jobs on the designated AI platforms, according to the data dependencies specified by the dataflow execution graph.

\section{System Evaluation}
\label{sec_evaluation}

This section evaluates the Gypscie system at two complementary levels: functional scope through qualitative comparison with representative AI systems, and operational behavior through an experimental assessment of its dataflow instantiation and scheduling functionality.
Section \ref{sec:qualitative_evaluation} provides a qualitative comparison of Gypscie with representative AI systems, focusing on model lifecycle management.
Section \ref{sec:exp-evaluation} gives our experimental evaluation.

\subsection{Qualitative Evaluation}
\label{sec:qualitative_evaluation}
Several systems have been proposed to support different aspects of the AI model lifecycle.
ML platforms such as MLflow \cite{chen_developments_2020} and Kubeflow \cite{korontanis_streamlining_2025} provide experiment tracking and pipeline orchestration capabilities, while tools like lakeFS \cite{lakefs_lakefscontrol_2026} focus on dataset versioning. Academic systems such as KeystoneML \cite{sparks_keystone_2017}, Helix \cite{xin_helix_2018}, and SystemDS \cite{baunsgaard_federated_2022} address scalable pipeline execution and optimization of iterative workflows, whereas ModelDB \cite{vartak_modeldb_2016} concentrates on ML model lifecycle management.
The systems included in this comparison were selected as representative industrial and academic approaches that cover functionalities closely related to Gypscie's scope, such as experiment tracking, dataflow orchestration, model lifecycle support, and dataset management. The comparison dimensions were defined from the AI artifact management capabilities discussed throughout this paper. In the tables, \emph{Yes} or \emph{Full} indicates direct support as a central capability, \emph{Partial} or \emph{Limited} indicates restricted or indirect support, and \emph{No} indicates that the capability is not explicitly provided.
In contrast, Gypscie employs an artifact-centric architecture that unifies the management of datasets, models, and dataflows within a single system. Data, metadata, and monitoring information are further consolidated into a knowledge graph, enabling explicit provenance modeling and rich semantic querying throughout the AI lifecycle. This design provides a cohesive, transparent, and explainable view of AI artifacts and their dependencies across heterogeneous AI platforms.

\begin{table}
\centering
\small
\caption{Gypscie versus industrial AI systems}
 \label{table:position_industry}
\renewcommand{\arraystretch}{1.15}
\begin{tabular}{|L{3.8cm}|C{1.8cm}|C{1.8cm}|C{1.8cm}|C{1.8cm}|}
\hline
 & Gypscie & MLflow & Kubeflow & LakeFS \\
 \hline
 Primary abstraction & Artifact graph & Experiments & Kubeflow pipelines & Data versions \\
 \hline
 Artifact-centric lifecycle model & Yes & Partial & No & Data-centric \\
 \hline
 Knowledge-graph representation & Yes & No & No & No \\
 \hline
 Explicit provenance modeling & Yes & Limited & Partial & Dataset lineage \\
 \hline
 Lifecycle coverage & Full & Partial & Full & Data-only \\
 \hline
 Dataflow modeling & Abs./Conc. & Declarative & Concrete & No \\
 \hline
 Model management & Yes & Yes & Ecosystem\textsuperscript{1} & No \\
 \hline
 Data versioning & Limited & Limited & External\textsuperscript{2} & Core \\
 \hline
 AI platforms & Multi\textsuperscript{3} & Local/Cloud & Kubernetes clusters & Storage\textsuperscript{4} \\
 \hline
 Cross-platform orchestration & Yes & Limited & No & Storage-level \\
 \hline
 Optimization of ML dataflows & Limited & No & Limited & No \\
 \hline
 Automatic selection of models & Limited & No & No & No \\
 \hline
 Integration with AI ecosystem & Yes & Yes & Yes & Yes \\
 \hline
 
\end{tabular}
\renewcommand{\arraystretch}{1}
\par\vspace{2pt}
\footnotesize{\textsuperscript{1} Via ecosystem tools. \textsuperscript{2} Requires external tools for data versioning. \textsuperscript{3} HPC, Spark clusters, and cloud environments. \textsuperscript{4} Storage-layer integration, e.g., S3 or GCS.}

\end{table}

Table \ref{table:position_industry} situates Gypscie relative to major AI systems in industry. Gypscie offers a comprehensive set of functionalities for managing the full lifecycle of AI artifacts. 
While MLflow is extensively used in industry and is also integrated within Gypscie, its primary focus on experiment tracking limits its ability to address higher-level requirements, such as knowledge graph integration and dataflow optimization.

Kubeflow is designed to support ML dataflows on Kubernetes infrastructures. While it provides mechanisms for orchestrating pipeline execution, it does not support knowledge graph representations and offers limited functionality for pipeline optimization. In addition, it does not directly address data management concerns and requires a Kubernetes cluster to execute pipelines.

Finally, lakeFS provides an effective solution for versioning datasets by adopting a Git-like version control model for data stored in data lakes. Although it offers powerful capabilities for dataset evolution and reproducibility, its scope is limited to the data layer and does not encompass the full lifecycle management of ML artifacts.
These comparisons highlight three distinctions of Gypscie with respect to representative industrial systems: broader lifecycle coverage, tighter integration between artifacts, provenance, and semantic querying, and support for cross-platform dataflow instantiation beyond a single execution platform.

\begin{table}
\caption{Gypscie versus academic AI systems}
\label{table:position_academy}
\centering
\small
\renewcommand{\arraystretch}{1.15}
\begin{tabular}{|L{3.2cm}|C{1.55cm}|C{1.55cm}|C{1.55cm}|C{1.55cm}|C{1.55cm}|}
\hline
& Gypscie & KeystoneML & Helix & ModelDB & SystemDS\\
\hline
Primary abstraction & Artifact graph & ML pipelines & ML pipelines & Models/runs & DS workflows \\
\hline
Artifact-centric lifecycle model & Yes & No & Partial & Yes & No \\
\hline
Knowledge-graph representation & Yes & No & No & No & No \\
\hline
Explicit provenance modeling & Full & Limited & Partial & Yes & Low\textsuperscript{1} \\
\hline
Lifecycle coverage & Full & Partial & Partial & Model-focused & Full \\
\hline
Dataflow modeling & Abs./Conc. & Pipeline DSL & Keystone DSL & Limited & Scripts\textsuperscript{2} \\
\hline
Model management & Yes & Limited & Limited & Core & Limited \\
\hline
Data versioning & Limited & Limited & Limited & Limited & Limited \\
\hline
Execution environments & Multi\textsuperscript{3} & Spark & Clusters & Local & Local/Spark \\
\hline
Cross-platform orchestration & Yes & Spark-centric & Limited & No & No \\
\hline
Optimization of ML dataflows & Limited & Yes & Yes & No & Yes \\
\hline
Automatic selection of models & Limited & No & No & No & Limited \\
\hline
Integration with ecosystem & Yes & Limited & Limited & Limited & API\textsuperscript{4} \\
\hline
\end{tabular}
\renewcommand{\arraystretch}{1}
\par\vspace{2pt}
\footnotesize{\textsuperscript{1} Low-granularity lineage. \textsuperscript{2} R and Python scripts. \textsuperscript{3} HPC, Spark clusters, and cloud environments. \textsuperscript{4} Java and Python API.}

\end{table}

Table \ref{table:position_academy} positions Gypscie with respect to important academic AI systems. KeystoneML and Helix were designed to optimize ML pipelines executed on Apache Spark. Similarly, Apache SystemDS extends the earlier Apache SystemML implementation \cite{boehm_systemml_2015}, incorporating lessons learned from data science workloads. The Apache SystemDS project focuses on providing a declarative abstraction layer for data science, combined with low-level runtime optimizations for efficient execution.

To support data locality and scalability, data can be distributed and managed across different storage infrastructures, including HPC distributed file systems, cloud storage services, or local storage. Finally, Gypscie supports applications built on top of its artifacts and metadata. To facilitate this integration, it provides a unified data view through a knowledge graph that enables structured access to artifacts and their relationships.
Relative to these academic systems, Gypscie is less focused on optimizing a single execution stack and more focused on combining artifact-centric lifecycle management, provenance, and cross-platform execution within the same architecture. This distinction is important because the goal of Gypscie is not only to execute pipelines efficiently, but also to manage the broader set of artifacts and relations involved in AI applications.

\subsection{Experimental Evaluation}
\label{sec:exp-evaluation}
A major functionality of Gypscie is the ability to define abstract dataflows that can be optimized and scheduled as concrete dataflows across multiple platforms.
In this section, we evaluate such functionality with experiments based on the use-case dataflow depicted in Figure \ref{fig_inference}. More specifically, the experiments pursue three complementary goals: to verify that the same abstract dataflow can be materialized on distinct AI platforms, to compare the performance behavior of two concrete instantiations of that dataflow, and to assess the impact of a simple rewriting strategy on execution cost.
The experiments therefore compare a Pandas-based execution on a server and a Spark-based execution on a cluster as two concrete instantiations of the same abstract dataflow. In this context, semantic consistency means that both instantiations preserve the same abstract operators, input-output structure, and expected processing semantics defined in the original dataflow specification.
In the rest of this section, we describe our experimental environment and methodology and analyze the experimental results.

\subsubsection{Experimental Environment.}
The experimental evaluation consists of a comparative analysis of execution time and RAM memory consumption between the two approaches (Pandas and Spark) to highlight performance and scalability differences. The experimental setup is as follows:
\begin{itemize}
\item 
The data are HDF-format files, with individual sizes ranging from 6 MB to 9 MB, generated every five minutes, totaling 288 files per day and approximately $5.2 \times 10^8$ records daily.
\item 
The experiments were conducted on a single dedicated host equipped with two Intel Xeon Silver 4216 processors, totaling two physical sockets, with a base frequency of 2.10 GHz. The system features 128 GB of DDR4 RAM (nominal frequency of 3200 MT/s, operating at 2400 MT/s) and a 1 TB NVMe SSD (Kioxia model KXG80ZNV1T02), used for both execution and data storage.
\item 
The distributed environment was provisioned using Docker containers on a Hadoop YARN-based cluster with one NameNode, one ResourceManager, two DataNodes, and two NodeManagers. Each NodeManager was configured with 25 vCores and approximately 51 GB of memory (52224 MB), with a maximum container allocation of 14 GB and a minimum of 1 GB.
\item 
Spark was configured using 8 executors, each with 4 cores and 6 GB of memory, along with an additional 4 GB of memory overhead. The driver was configured with 2 cores and 6 GB of memory (plus 2 GB of overhead), and the shuffle parallelism was set to 96 partitions.
\item Gypscie was deployed using Docker containers, ensuring the isolation, reproducibility, and consistency of the experimental environment across all executions.
\end{itemize}

\subsubsection{Experimental Methodology.}
The experiments follow a controlled methodology, in which the host is kept under exclusive use throughout execution to minimize external interference and ensure measurement consistency. Seven data samples were considered, consisting of ${10, 20, \ldots, 70}$ weather radar files, representing different input sizes. Each sample was processed five times using both dataflow implementation approaches (Pandas and Spark), and the average and standard deviation of the execution time were computed for each configuration.

For Pandas, the behavior of RAM memory consumption was evaluated as data size increases. This analysis is intended to characterize a concrete bottleneck of the Pandas-based instantiation in the studied scenario, rather than to provide a directly symmetric memory comparison between the two execution technologies.
We did not perform such analysis for Spark, as it provides built-in mechanisms for memory management and optimization. In addition to the main comparison between Pandas and Spark, we also include an illustrative optimization case study. The sample composed of 70 files was reprocessed using an optimized version of the dataflow, in which the filtering dataflow graph node was moved earlier in the dataflow to reduce the size of data processed earlier, thereby improving efficiency in terms of both memory usage and execution time.

We implemented the functions for each dataflow graph node operator in Python, using the native functionality of the respective technologies (Pandas and Spark). Each implemented function adheres to the operations defined in the abstract dataflow and to the specified input and output data structures, ensuring semantic consistency across executions and alignment with the abstract specification of the use-case dataflow.

\subsubsection{Experimental Results.}
In this section, we first compare the execution times of the use-case dataflow implemented with the Pandas and Spark approaches, then analyze RAM memory consumption for the Pandas approach, and finally discuss how a simple rewriting strategy affects the cost of one concrete materialization.

\begin{figure}[ht]
    \centering
    \includegraphics[width=0.8\linewidth]{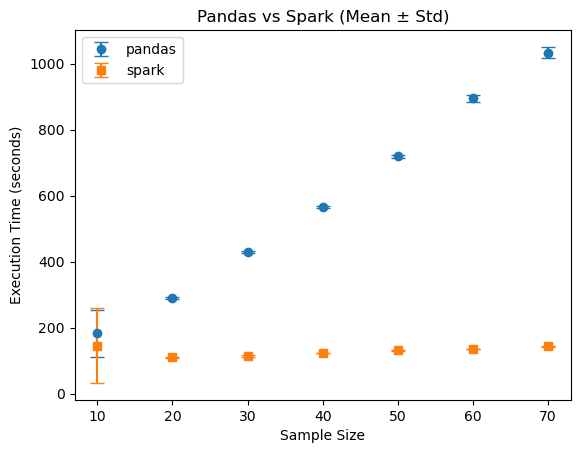}
    \caption{Average execution time and standard deviation for Pandas and Spark as a function of the sample size, i.e., the number of radar files with volumes ranging from 6 MB to 9 MB.}
    \label{fig:execution_time}
    \end{figure}

Figure~\ref{fig:execution_time}
shows the average execution time and standard deviation for Pandas and Spark as the data size increases, with significant differences in performance behavior. 
For the data sample of size 10, both approaches exhibit higher variability in execution time, reflected by a larger standard deviation. Such behavior is explained by the presence of an initial overhead associated with the first execution, related to the setup of the Conda environment, which does not occur in subsequent runs and therefore does not affect the remaining samples.

From the following data samples onward, execution times become stable, with low variability, indicating consistency in the measurements. In terms of scalability, the Pandas approach shows an approximately linear increase in execution time as the sample size grows, highlighting its limitations in handling larger data sizes. In contrast, the Spark approach maintains relatively constant execution times across different samples, demonstrating superior scalability and efficiency in distributed processing. These results reinforce the suitability of each technology for different data sizes and performance requirements.
From the standpoint of Gypscie, this result is important because it shows that cross-platform instantiation is not merely a portability feature. Different concrete instantiations of the same abstract dataflow can exhibit markedly different operational behavior, which justifies platform-aware scheduling.

\begin{figure}[ht]
    \centering
    \includegraphics[width=1.0\linewidth]{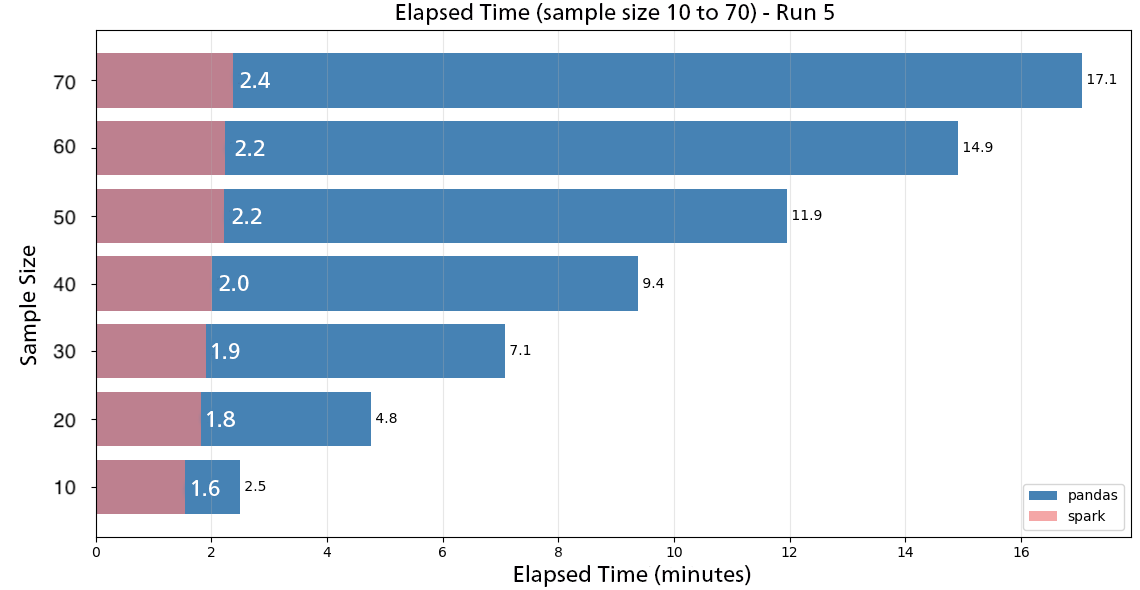}
    \caption{Total execution time (in minutes) per sample, considering the fifth run, for Pandas and Spark. The sample size corresponds to the number of radar files with volumes ranging from 6 MB to 9 MB.}
    \label{fig:total_execution_time}
    \end{figure}

Figure~\ref{fig:total_execution_time} provides a complementary view of the results by considering only the fifth execution of each sample, since execution times become consistent after the initial stabilization phase. The analysis therefore focuses on the approximate absolute execution time, expressed in minutes, providing a direct comparison of computational cost between the approaches.

These results further highlight the scalability difference between the two approaches. While Pandas is suitable for smaller data sizes, Spark delivers superior and more stable performance in larger-scale scenarios, even when considering absolute execution time.

\begin{figure}[ht]
    \centering
    \includegraphics[width=1.0\linewidth]{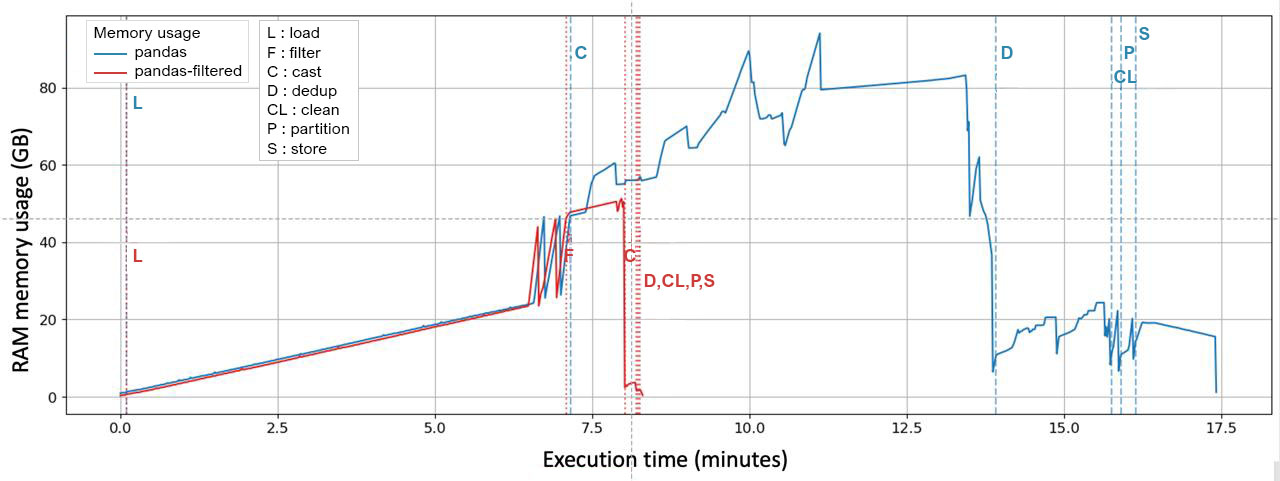}
    \caption{Comparison of RAM usage over time between the original and optimized dataflows, considering a sample of 70 files processed using Pandas.}
    \label{fig:memory_usage_pandas_70_filtered}
    \end{figure}

The analysis of RAM memory consumption for the Pandas approach reveals a behavior directly proportional to the size of processed data. As shown in Figure \ref{fig:memory_usage_pandas_70_filtered},
peak memory usage increases from approximately 13.37 GB for the sample with 10 files to 93.31 GB for the sample with 70 files, indicating an almost linear growth with respect to input size. This behavior highlights the limitations of this approach in larger-scale scenarios.

\begin{figure}[ht]
    \centering
    \includegraphics[width=1.0\linewidth]{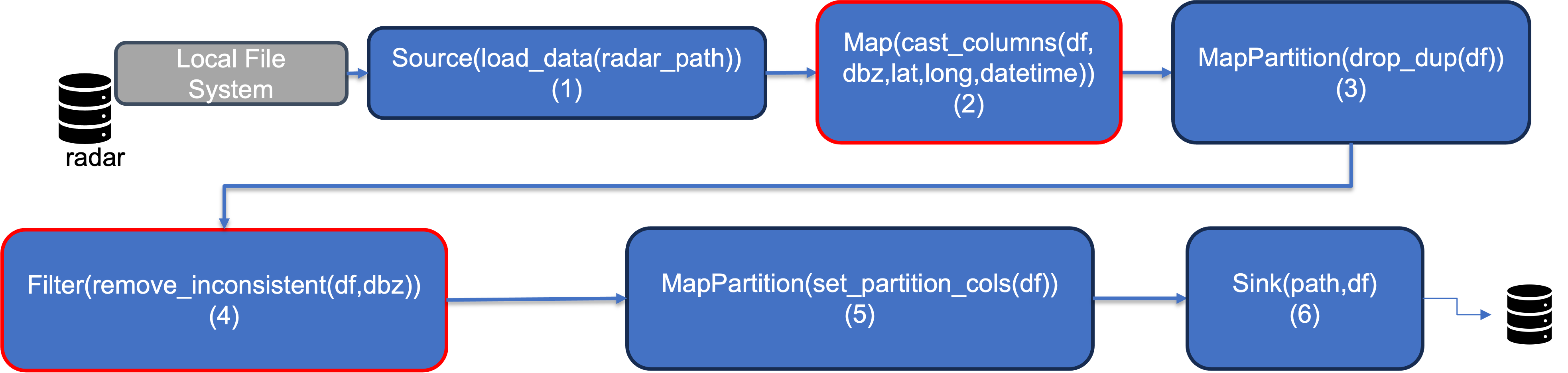}
    \caption{Pre-processing dataflow for radar data. The Filter operator selects measurements from a radar dataset with high precipitation volume values, ($dbz \geq 50$). When placed before the \textit{Map(cast\_columns)} dataflow operation, it avoids the extra cost of casting values of tuples that will be later filtered out.}
    \label{fig:pushdown-dataflow}
    \end{figure}

Taking into account the memory consumption and the total execution time for both implementations of the same abstract dataflow, the relevance of capturing retrospective provenance information about dataflow runs, and using these data to feed Gypscie's optimized scheduling process becomes evident.
This observation further reinforces the need for Gypscie to reason about execution history when deciding how to initialize and schedule dataflows, since the same logical specification may lead to very different resource profiles depending on the chosen execution technology.
In this second experiment, we ran the pre-processing radar dataflow depicted in Figure \ref{fig:pushdown-dataflow}.

Figure~\ref{fig:memory_usage_pandas_70_filtered} provides a detailed temporal profile of memory usage for the dataflow in Figure \ref{fig:pushdown-dataflow}, using Pandas, for the sample of size 70 (execution 5), enabling a finer-grained analysis across dataflow stages. The cast stage is responsible for the highest memory consumption, reaching peak values during execution and accounting for a significant portion of the total runtime. In contrast, subsequent stages, such as dedup and store, exhibit sharp reductions in memory usage, associated with the release of intermediate data structures.

The Gypscie optimizer evaluates a number of heuristics to produce an efficient concrete dataflow. In the example use case, the memory consumption and processing time introduced by the cast function can be optimized. Looking more closely at the cast function code, we observe that three columns undergo different casting transformations. For the dbz column, i.e., radar reflectivity, the resulting column values are used later in the dataflow for filtering, retaining only values above a certain threshold that identify strong rainfall.
However, the dbz casting is hidden within the cast function, which makes it invisible to the Gypscie optimizer.
A simple solution is to make the dbz cast function a node in the dataflow. This would allow the Gypscie optimizer to apply the well-known filter pushdown approach,
thus reducing the size of processed data at an early stage and, consequently, improving overall system performance.

We applied this solution in an alternative, optimized dataflow that moves the filter stage, originally positioned after more costly transformations, to precede the cast stage. Such reordering enables an early reduction in the size of data processed in subsequent stages, positively impacting both memory consumption and total execution time.
This optimization case study is particularly relevant for Gypscie because it shows that the abstract dataflow representation is useful not only for execution across platforms, but also for exposing optimization opportunities that would be harder to identify and exploit in a less structured dataflow specification.

Figure~\ref{fig:memory_usage_pandas_70_filtered} compares memory usage between the original Pandas dataflow and its optimized version, considering the sample of 70 files (execution 5). 
In the original version, memory usage peaks above 90 GB during the cast stage, whereas in the optimized version this value is reduced to approximately 50 GB, representing a substantial decrease in peak resource consumption. In addition, total execution time is significantly reduced, as computationally intensive operations are applied to a smaller subset of data, thus improving overall efficiency.

    In summary, the results demonstrate that Gypscie is able to materialize concrete dataflows from an abstract definition while preserving semantic coherence across different execution platforms. The experimental evaluation revealed significant differences, highlighting the limitations of the Pandas approach in terms of scalability, both in execution time and memory consumption, in contrast to the more stable and efficient behavior of Spark in larger-scale scenarios. These findings show that Gypscie can represent a dataflow at a high level, instantiate it on distinct AI platforms while preserving its intended semantics, and benefit from optimization strategies that improve the operational cost of concrete executions.

\section{Related Work}
\label{sec_related-work}

Spurred by the growing adoption of AI across various applications, numerous new systems have been proposed to support the entire AI model lifecycle. Unlike traditional software engineering, the development of AI applications is more iterative and exploratory, yielding a variety of artifacts, including datasets, models, feature sets, model parameters, hyperparameters, metrics, software code, and dataflows. The objective of these systems is to enable explainability, reproducibility, traceability, and comparability of AI executions by supporting the storage, management, and reuse of these artifacts.

A systematic literature review of more than 60 such systems, referred to as ML artifact management systems \cite{schlegel_management_2023}, reveals that there is no precise functional scope, thereby making comparisons between systems difficult. Some systems focus on a single class of artifact, while others combine artifact management with development and execution support. To position Gypscie more precisely, we discuss the literature along four complementary axes.

First, several systems address AI artifact management across substantial portions of the lifecycle. The most comprehensive systems are the major ML-as-a-Service (MLaaS) offerings, such as Azure ML, Amazon SageMaker, and Google Vertex AI, but they are tightly coupled to specific AI platforms or provider ecosystems. Open-source platforms such as MLflow \cite{chen_developments_2020}, ClearML \cite{clearml_clearml_2025}, and Hopsworks \cite{ismail_hopsworks_2017,hopsworks_hopsworks_2025} provide reusable services for tracking, storing, and operating on ML artifacts, while emphasizing different concerns such as experiment tracking, operational support, or metadata and data infrastructure management. These systems provide important lifecycle support, yet they generally do not expose a unified semantic representation that integrates artifacts, metadata, provenance, and execution relationships. In particular, they offer powerful lifecycle services, but they do not explicitly treat heterogeneous AI artifacts and their interdependencies as part of a single logic-based integration view.

Second, a substantial body of work focuses on workflows, pipelines, and execution environments. Kubeflow \cite{korontanis_streamlining_2025,kubeflow_kubeflow_2025}, TensorFlow Extended with ML Metadata \cite{baylor_tfx_2017}, Apache SystemDS \cite{boehm_systemds_2020}, and ArangoML Pipeline \cite{arangodb_arangoml_2025} emphasize orchestration, modularity, lineage, and reproducibility in data processing, model training, and deployment pipelines. Academic systems such as KeystoneML \cite{sparks_keystone_2017}, Helix \cite{xin_helix_2018}, and LifeSWS \cite{akbarinia_life_2023} further highlight optimization, scalability, debugging, and workflow reuse, especially in scientific settings. KeystoneML and Helix, in particular, explore optimization of ML pipelines executed on distributed infrastructures, whereas LifeSWS extends workflow-oriented support toward scientific applications by making workflow artifacts easier to search, debug, and parallelize \cite{akbarinia_life_2023}. However, these systems are primarily execution-oriented: they represent workflows well, but they typically provide limited support for a unified, artifact-centric view spanning datasets, models, provenance, and domain semantics across the entire lifecycle. Their main strength lies in execution planning and pipeline management rather than in offering an integrated representation of all AI artifacts manipulated by an application.

Third, other systems provide deeper management for specific artifact classes. Experiment management tools such as Weights \& Biases \cite{weights__biases_ai_2025}, Sacred \cite{idsia_sacred_2025}, and Guild AI \cite{guild_ai_guild_2025} strengthen comparability across training runs, but they remain centered on run metadata rather than end-to-end lifecycle integration. Model-oriented systems such as ModelDB \cite{vartak_modeldb_2016}, ModelHub \cite{miao_modelhub_2017}, and MLModelCI \cite{zhang_mlmodelci_2020} treat trained models and their metadata as first-class entities, supporting versioning, deployment, and monitoring. Data- and feature-oriented systems such as Feast \cite{feast_feast_2025}, lakeFS \cite{lakefs_lakefscontrol_2026}, Pachyderm \cite{novella_container-based_2019}, and ProvLake \cite{souza_workflow_2022} emphasize storage, versioning, and provenance tracking for datasets and features. These systems illustrate that the literature offers mature solutions for isolated dimensions of the problem: experiment tracking, model governance, or dataset and feature management. However, each remains specialized in a subset of lifecycle stages or artifact types. As a result, an application that relies on several of these capabilities still depends on external integration effort to connect models, datasets, transformations, provenance, and execution context in a coherent manner.

Finally, recent work on LLM-related artifacts reinforces the need for stronger provenance, transparency, auditability, and lifecycle management operations. RAGOps \cite{xu_ragops_2025}, for example, targets the continuous management of retrieval-augmented generation pipelines in production environments. Synthetic Artifact Auditing \cite{wu_synthetic_2025} focuses on detecting downstream artifacts derived from LLM-generated synthetic data, exposing risks related to traceability, bias, and hallucination. Although these proposals address emerging challenges specific to LLM-based systems, they also confirm a broader trend: AI applications increasingly require integrated management of heterogeneous artifacts, their transformations, and their operational context. In this sense, LLM-oriented work does not define a separate problem from the rest of the AI lifecycle. Rather, it amplifies existing demands for integrated provenance, artifact correlation, and support for evolving execution pipelines.

In contrast, Gypscie is a high-level, generic system that unifies AI artifact management across the lifecycle using a knowledge graph, supports interoperability with diverse AI frameworks, and provides flexible integration with heterogeneous AI platforms through a high-level dataflow language.

\label{sec_conclusion}

Gypscie is a cross-platform AI artifact management system that unifies artifact management across the entire AI model lifecycle. In this paper, we presented the architecture of Gypscie and its set of services for managing and utilizing datasets, models, functions, dataflows, and provenance information across ML and LLM applications.

Gypscie's architecture builds on advances in web-oriented architectures, containers, and distributed and parallel data management. Gypscie offers both a web interface that provides user-friendly access to services and an API that enables seamless integration with complex applications. To reduce the complexity of developing and deploying AI applications, Gypscie provides a unified view of all artifacts. This view is implemented through a knowledge graph that captures application semantics and a rule-based query language, which enables the retrieval of artifact metadata, domain data, predictions, and provenance information about artifacts and service execution.

Model lifecycle activities are expressed as high-level dataflows, which can be scheduled to run on platforms such as servers, clouds, or supercomputers. We introduce a high-level dataflow language to represent these dataflows as DAGs of data transformations. Dataflow processing includes optimization, scheduling, and execution, thereby improving efficiency and resource utilization.

Gypscie captures and manages provenance information about artifacts and their transformations, thereby supporting explainability, reproducibility, and traceability across diverse application domains.

We provide a comprehensive evaluation of the Gypscie system. A qualitative comparison with representative AI systems shows that Gypscie supports a broader range of functionalities across the AI artifact lifecycle. It provides detailed provenance information for managed artifacts, enabling explainability and reproducibility throughout the entire lifecycle. Its operator-based dataflows also facilitate automatic optimization and scheduling, allowing them to scale from desktop environments to large infrastructures, including HPC AI platforms. To the best of our knowledge, Gypscie is the only system that supports a unified data and metadata view as a knowledge graph. These qualitative results show that Gypscie supports integrated AI artifact management by combining semantic querying, provenance support, and cross-platform execution within a single architecture.

The experimental evaluation focuses on a key functionality of Gypscie: the ability to optimize and schedule abstract dataflows into concrete executions across multiple platforms. This functionality is assessed through experiments based on a use-case dataflow deployed on two platforms: one capable of running a Python application on a Pandas server and another supporting an Apache Spark dataflow on a cluster. The results demonstrate that Gypscie can successfully materialize concrete dataflows from an abstract specification while preserving semantic consistency across both AI platforms.
The experiments revealed significant performance differences between the two platforms, highlighting the scalability limitations of the Pandas-based approach in terms of execution time and memory consumption. In contrast, Spark exhibited more stable and efficient behavior in large-scale scenarios.
The results also show that the ordering of dataflow operators has a direct impact on performance. Substantial improvements can be achieved through simple restructuring strategies, such as applying filtering operations earlier in the dataflow.

In this context, the experimental results reinforce the practical relevance of Gypscie's cross-platform dataflow instantiation and scheduling capabilities, while also highlighting opportunities for further automation. They show that Gypscie can represent dataflows at a high level, instantiate them across distinct AI platforms while preserving their intended semantics, and benefit from optimization strategies that reduce execution cost. Gypscie thus emerges as a promising and evolving system, with the potential to incorporate automated dataflow optimization mechanisms that can transparently exploit such opportunities for performance improvement.

As future work, we plan to further investigate these optimization strategies, aiming not only to improve computational efficiency but also to enhance the system's adaptability to diverse large-scale data processing scenarios.

\section*{Acknowledgement}

This work was carried out within the context of the Dinizia associated team, a collaboration between Inria and Brazil, and Fabio Porto's International Chair at Inria. It was supported by ANP Cooperation No. 0050.0122040.22.9, the Brazilian National Council for Scientific and Technological Development (CNPq), and the AI Institute at LNCC.

\bibliographystyle{acm}
\bibliography{references}

\end{document}